\documentclass[letterpaper, 10 pt, conference]{ieeeconf}  

\IEEEoverridecommandlockouts                              
\usepackage{todonotes}
\usepackage{graphicx} 
\usepackage{times} 
\usepackage{amsmath} 
\usepackage{amssymb}  
\usepackage{todonotes}
\usepackage{verbatim} 
\usepackage{cite}
\usepackage{subfigure}
\usepackage[hyphens]{url}
\usepackage[ruled,vlined]{algorithm2e}
\usepackage{float}
\usepackage{tabulary}
\usepackage{tabularx}
\usepackage{lipsum,amsmath,multicol}
\usepackage{tensor}
\usepackage{multirow}
\usepackage{booktabs}

\def\fps@figure{htp}
\def\fps@table{htp}

\newcommand{\bi}{\begin{itemize}}
\newcommand{\ei}{\end{itemize}}

\newcommand{\bfig}{\begin{figure}}
\newcommand{\efig}{\end{figure}}

\newcommand{\benum}{\begin{enumerate}}
\newcommand{\eenum}{\end{enumerate}}

\newcommand{\be}{\begin{equation}}
\newcommand{\ee}{\end{equation}}

\newcommand{\ba}{\begin{eqnarray}}
\newcommand{\ea}{\end{eqnarray}}

\newcommand{\etal}{{et al.}}

\newcommand{\unit}[1]{\mbox{$\rm \,#1$}}

\usepackage{color}

\definecolor{CommentRed}{rgb}{0.7,0,0}
\definecolor{CommentBlue}{rgb}{0,0,0.7}
\definecolor{CommentDG}{rgb}{0,0.6,0}

\newcommand{\coordi}[1]{$\{\mathcal{#1}\}$}

\newcommand{\stxt}[1]{\mbox{\tiny $\text{#1}$}}
\newcommand{\bs}[1]{\boldsymbol{#1}}

\makeatletter
\newenvironment{myalign*}{%
  \setlength{\mathindent}{0pt}%
  \setlength{\abovedisplayskip}{-\baselineskip}%
  \setlength{\abovedisplayshortskip}{\abovedisplayskip}%
  \start@align\@ne\st@rredtrue\m@ne
}%
{\endalign}
\makeatother

\title{\LARGE \bf
Dynamic System Identification, and Control for a cost effective open-source VTOL MAV
}

\author{$\text{Inkyu Sa}^{1}$, $\text{Mina Kamel}^{1}$, $\text{Raghav Khanna}^{1}$, $\text{Marija Popovi\'{c}}^{1}$, $\text{Juan Nieto}^{1}$, and $\text{Roland Siegwart}^{1}$\thanks{${}^1$ Autonomous Systems Lab., Department of Mechanical and Process Engineering, ETH Zurich, Zurich, Switzerland. \texttt{inkyu.sa@mavt.ethz.ch}}
\thanks{This project has received funding from the European Union's Horizon 2020 research and innovation programme under grant agreement No 644227 and from the Swiss State Secretariat for Education, Research and Innovation (SERI) under contract number 15.0029.}
}

\begin{document}

\maketitle

\thispagestyle{empty}
\pagestyle{empty}


\begin{abstract}
This paper describes dynamic system identification, and full control of a cost-effective vertical take-off and landing (VTOL) multi-rotor micro-aerial vehicle (MAV) --- DJI Matrice 100. The dynamics of the vehicle and autopilot controllers are identified using only a built-in IMU and utilized to design a subsequent model predictive controller (MPC). Experimental results for the control performance are evaluated using a motion capture system while performing hover, step responses, and trajectory following tasks in the present of external wind disturbances. We achieve root-mean-square (RMS) errors between the reference and actual trajectory of x=0.021\unit{m}, y=0.016\unit{m}, z=0.029\unit{m}, roll=0.392\unit{{}^\circ}, pitch=0.618\unit{{}^\circ}, and yaw=1.087\unit{{}^\circ} while performing hover. This paper also conveys the insights we have gained about the platform and returned to the community through open-source code, and documentation.
\end{abstract}

\section{INTRODUCTION}
\label{sec:intro}
VTOL MAV platforms are rotorcraft air vehicles that use counter-rotating rotors to generate thrust and rotational forces. These vehicles have become a very popular research and commercial platform during the past decade. The wide variety of ready-to-fly platforms today is proof that they are being utilized for real-world aerial tasks such as indoor and outdoor inspection, aerial photography, cinematography, and environmental survey and monitoring for agricultural applications. The performance of these vehicles has also shown steady improvement over time in terms of flight time, payloads, and safety-related smart-features. For example, products from the leading drone manufacture, DJI, can assist a pilot with a stereo vision-based positioning system that provides more stable, easier, and safer flight. Total flight time is also getting extending with recent advances in battery, integrated circuit, and material technologies.

However, while these features and performance are sufficient for a manual pilot for the collection of imagery or GPS-based autonomous navigation, to achieve the tasks above. 
It is challenging to adapt these commercial platforms for robotic tasks such as obstacle avoidance and path planning \cite{burri2015real,Nuske:2015aa,Yang-:2015aa}, landing on a moving platform\cite{Lee:2012aa}, object picking\cite{mellinger2011design}, and precision agriculture\cite{Zhang:2012aa}.

To achieve these tasks, an accurate dynamics model, a low-latency and precise state estimator, and a high-performance controller are required. Ascending Technologies provides excellent research grade VTOL MAV platforms \cite{Achtelik:2011fk, Weiss:2011aa} dedicated to advanced aerial robotic applications. There is also a well-explained software development kit (SDK) and abundant scientific resources such as self-contained documentation. These platforms are ideal for developing aerial robots. However, their expensive cost is a major hurdle for many robotic researchers and replacing parts in a case of a crash (which can occur in the early development stage) is time-consuming and expensive due to the limited number of retail shops.

\begin{figure}
\begin{center}
\includegraphics[width=\columnwidth]{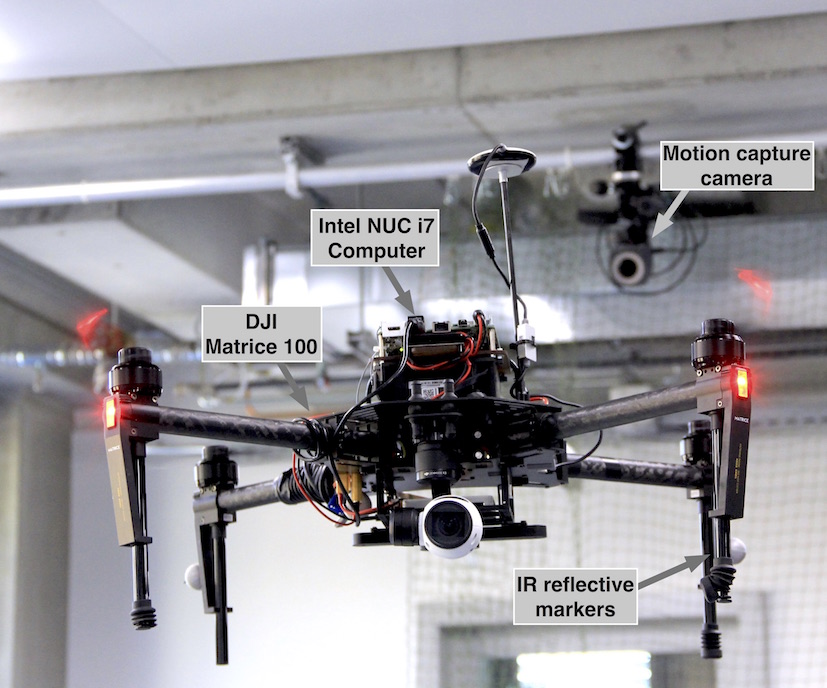}
\end{center}
    \vspace{-5mm}
    \caption{DJI Matrice 100 VTOL MAV quadrotor platform. An onboard computer is mounted, and the external motion capture device provides accurate position, velocity, and orientation measurements at 100\unit{Hz} for Model Predictive Controller (MPC).}
    \label{fig:M100}
    \vspace{-6mm}
\end{figure}

For VTOL MAVs to become more pervasive, they must become lower in cost and their parts easier to replace. DJI, a \$8 billion valuation drone manufacturer, has recently released the M-series of VTOL MAV platforms, shown in Fig.~\ref{fig:M100}, for developers with a SDK and documentation. Developers can now access sensor data such as IMU and barometers and send command data to the low-level attitude controller. In 2015, DJI has around 70\% drone market share that implies these platforms are affordable, and it is easy to order parts from local retail stores with short delivery spans. There are, however, also difficulties in using the M-series platform for robotics applications. Although DJI provides SDK and documentation, there is still a lack of essential scientific resources such as attitude dynamics and the structure of underlying autopilot controller. This information is critical for the subsequent controller such as MPC.

In this paper, we address these gaps by performing system identification both in simulated and experimental environments using only the built-in onboard IMU. Researchers can perform their system identification with the provided documentation and open-source code. To the authors' knowledge, this is the first attempt to identify systems of DJI M-series VTOL MAV platform that can be utilized for robotic applications.  

The contributions of this system paper are:
\begin{itemize}
\item Presenting full dynamics system identification using only onboard IMU that is used by a subsequent MPC-based position controller.
\item Sharing knowledge that can be essential for developing an M-series VTOL MAV platform.
\item Delivering software packages including modified SDK, linear MPC, and system identification tools and their documentation to the community. \begin{center}\texttt{http://goo.gl/lXRnU8}\end{center}
\end{itemize}

The benefit of this paper would be that the proposed techniques can be directly applied to other products of DJI such as Phantom-series and Inspire for fully autonomous maneuvers through their mobile SDK.

The remainder of this paper is structured as follows. Section \ref{sec:background} introduces state-of-the-art work on MAV system identification and control, and Section \ref{sec:methodologies} describes the specification of the vehicle, system identification and control strategies. We present our experimental results in Section \ref{sec:results}, and conclude in Section \ref{sec:conclusion}.
\vspace{-5pt}
\section{Related Work/Background}\label{sec:background}

VTOL MAVs' popularity is gaining momentum both industry and research field. It is necessary to identify their underlying dynamics system and behavior of attitude controllers to achieve good control performance. For a common quadrotor, such as the DJI Matrice 100, the rigid vehicle dynamics are well-known \cite{bouabdallah2007design} and can be modeled as a non-linear system with individual rotors attached to a rigid airframe, taking into account of drag force and blade flapping~\cite{mahony2012multirotor}. However, the identification of attitude controllers is often a non-trivial task for such consumer products due to the lack of scientific resources.

There are several system identification techniques to estimate dynamic model parameters. System identification methods can be three-fold: (i) offline, (ii) online, and (iii) batch techniques. Traditionally, parameter estimation has been performed offline using complete measurement data obtained from a physical test bed and CAD models~\cite{Pounds:2009fk, hoffmann2008quadrotor}. Such pioneer offline methods significantly contributed to the VTOL MAV community in the early development stage. However, the methods require high-precision measurement equipments which may be unsuitable to the small-scale consumer products\cite{Pounds:2010aa}. Alternatively, online system identification involves applying recursive estimation to real-time flight data. Chowdhary \etal \cite{chowdhary2010aerodynamic} study solutions in this category using different Kalman Filter-based approaches. In batch processing, a linear least-squares method is used to estimate parameters from recorded flight data \cite{Sa:2012ICRA,tischler2006aircraft}. Recently, Burri \etal \cite{burri2016maximum} demonstrated a method for identification of the dominant dynamic parameters of a VTOL MAV using Maximum Likelihood approach. In this study, they demonstrated accurate estimation of moments of inertia, the center of gravity in IMU frame, and aerodynamics parameters such as rotor thrust, rotor moment, and drag coefficients. However, a  detailed physical model of a VTOL MAV is required, and the underlying attitude controller's behavior was not identified. We also follow a batch-based approach to determine the dynamic parameters of the DJI Matrice 100 from short manual pilots. This allows us to obtain the parameters necessary for MPC using only the onboard IMU and without applying any restrictive simplifying assumptions.

Given the identified dynamics model, we use a high-performance state-of-the-art Model Predictive Control (MPC)\cite{kamelmpc2016} for horizontal position control.
\vspace{-3pt}
\section{Matrice 100 VTOL MAV platform}
\label{sec:methodologies}
In this section, we present overviews for the hardware platform, software development toolkit, and address attitude dynamics, and control strategy.
\vspace{-5pt}
\subsection{Coordinate systems definition}
\begin{figure}
\includegraphics[width=\columnwidth]{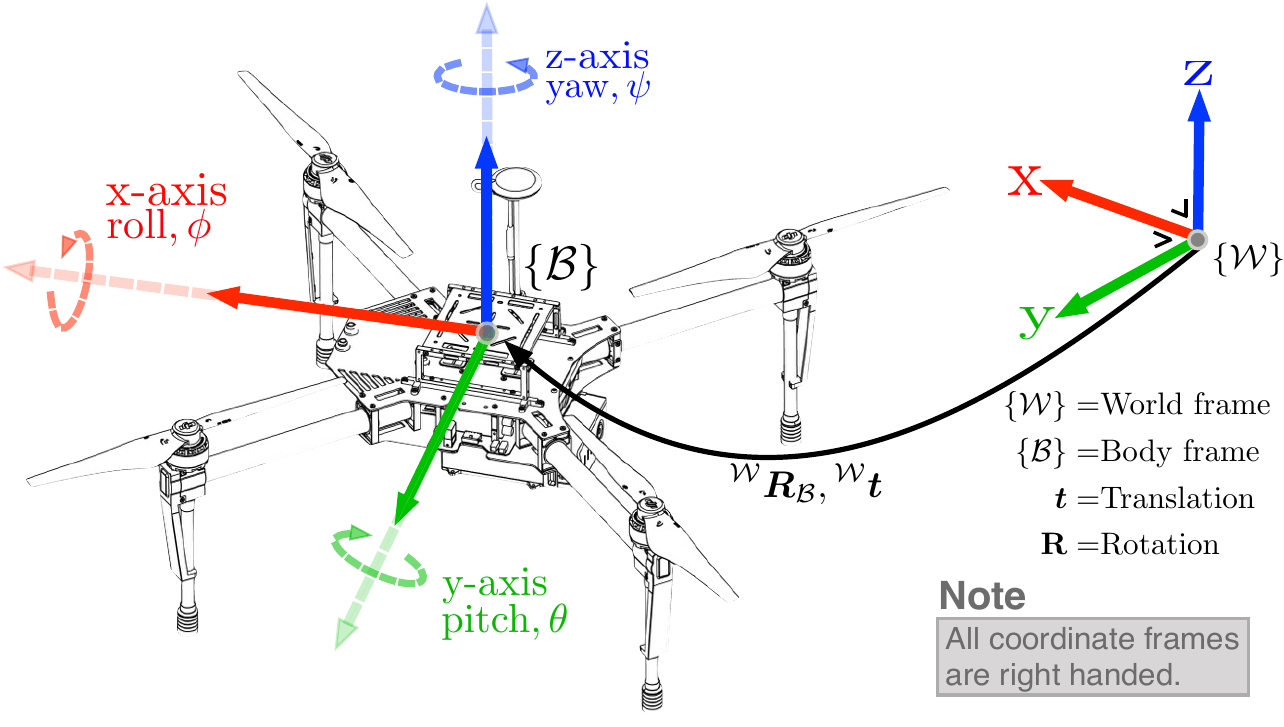}
    \vspace{-5mm}
\caption[Caption for LOF]{Coordinate systems definition\protect\footnotemark. ${}^\mathcal{W}\boldsymbol{t}$ is the 3$\times$1 translation vector w.r.t  \coordi{W}. ${}^\mathcal{W}\boldsymbol{R}_{\mathcal{B}}$ is 3$\times$3 matrix and rotates a vector defined w.r.t \coordi{B} to a vector w.r.t \coordi{W}.}
\label{fig:coordi}
    \vspace{-5mm}
\end{figure}
\footnotetext{Image source from http://goo.gl/7NsbmG}

We define 2 right-handed frames following standard Robot Operating System (ROS) convention: world $\{\mathcal{W}\}$ and body $\{\mathcal{B}\}$ shown in Fig \ref{fig:coordi}. x-axis in $\{\mathcal{B}\}$ indicates forward direction for the vehicle, y-axis is left, and z-axis it up. We use Euler angles; roll ($\phi$), pitch ($\theta$), and yaw ($\psi$) about x, y, z axes respectively for the RMS error calculation and visualization purposes. Quaternions are utilized for any computational processes. Note that Matrice 100 has North-East-Down coordinate and it is rotated by $\boldsymbol{R}=\boldsymbol{R}_{\mbox{\tiny x}}(\pi)\cdot\boldsymbol{R}_{\mbox{\tiny M100}}$ where $\boldsymbol{R}_{\mbox{\tiny M100}}$ is angle measurement from onboard IMU and $\boldsymbol{R}_{\mbox{\tiny x}}(\pi)$ is rotating $\pi$ along x-axis. The same operation is applied to acceleration and angular velocity measurements. The defined coordinate systems and notations are used over the rest of paper.
\vspace{-10pt}
\subsection{Hardware}
The general hardware specification of the M100 is well documented, and its CAD model is available from the official site, this section highlights our findings. 
The vehicle is a quadrotor and has 650\unit{mm} diagonal length. It uses N1 flight controller, but the information regarding the device is not disclosed to the public.

\subsubsection{Sensors update rates}
The variety of sensing data can be accessed using SDK through serial communication, such as IMU, GPS, barometer, magnetometer, and ultrasound measurements. A user can also configure update rate for the sensor up to 100\unit{Hz}. Importantly, the default sensor update set as 50\unit{Hz} with 230,400\unit{bps} baudrate. In order to measure at 100\unit{Hz}, the baudrate needs to be changed to 921,600\unit{bps} to avoid possible packet loss.

\subsubsection{Stick inputs}
The common transmitter inputs are pitch, roll angles (\unit{rad}), yaw rate (\unit{rad/s}), and thrust (\unit{N}). However, the vertical stick input of the platform is velocity (\unit{m/s}) that permits easier and safer manual flight. To address this different, we use a classic PID vertical position controller alongside linear MPC horizontal position controller.

The autopilot compensates total thrust variation in the translational maneuver. There are three control mode named: `F', `A', and `P'. The `F' mode allows external control inputs such as serial commands. `A' and `P' indicate attitude and position control modes. `A' mode should be selected to correctly track a reference command since `P' mode runs the manufacture's velocity controller that degrades the tracking performance.

\subsubsection{Auto-trim compensation}
There is no trim button on the provided transmitter; instead, the N1 autopilot has auto-trim functionality that balances attitude by estimating horizontal velocity. This feature allows easier and safer manual piloting but introduces a constant offset position error for controlling. More precisely, we can use Virtual RC commands to send control commands which have the minimum value of (1024-660) and the maximum (1024+660). In the ideal case (simulator), 1024 is the neutral value of Virtual RC command, and the quadrotor should hover at a position. In practice, however, the balancing point can be slightly higher or lower than the neutral value due to an unbalanced inertia of the vehicle. This needs to be considered and appropriately compensated. We estimate the balancing point where the vehicle's motion is minimum (hovering). It turns out that roll and pitch are balanced around 1080 and 998 respectively. We then adjust the neutral position to the estimated balancing point. If there is a change in an inertial moment (e.g., mounting a new device or changing the battery position), the balancing position has to be updated.

\subsubsection{Hardware-in-loop simulator}
The manufacturer provides the hardware-in-loop simulator that communicates with the autopilot on Windows and Mac. The primary functions of the simulator are receiving input commands from the transmitter and publishing the vehicle state. This tool is useful for debugging and algorithms testing without actual flights, and the identified dynamics system are surprisingly close to the real aircraft. We will cover this in the \ref{sec:dynamics} section.

\subsubsection{Dead zone recovery}\label{sec:deadZone}
Another interesting aspect of the autopilot is the presence of dead zone that is the small range close to the neutral value. Within this zone, the autopilot ignores the input commands, and no API is supported to set this. This function is also useful for a manual pilot since the vehicle should not react to small inputs yielded by tremor of hands but degrades the performance of the controller. We determine this by sweeping control commands around dead zone area and detecting the control inputs when any motion is generated (i.e., horizontal and vertical velocities changes). Although this task is difficult with a real VTOL platform due to its fast and naturally unstable dynamics, but we use the hardware-in-loop simulator that enables to receive input commands from the transmitter. In the simulator, the vehicle can hover at the same position without control inputs. The dead zone for pitch and roll is $\pm$19.8, $\pm$30.5 for height, and $\pm$30.5 for yaw rate in Virtual RC scale. If the commands are within those ranges, we set them as the maximum/minimum dead zone values (e.g., $\pm$19.8 for pitch and roll).

\subsection{Software Development Kit, SDK}
The SDK enables access to most functionalities and supports cross-platform development environments such as Robot Operating System (ROS), Android, and iOS. We use the onboard SDK with the ROS wrapper in this paper, but there is a fundamental issue for sending control commands with this protocol. The manufacturer uses \texttt{ROS services} to send commands that is strongly not recommended\footnote{http://wiki.ros.org/ROS/Patterns/Communication}. It is a blocking call that should be used for triggering signals or quick calculations. If data transaction (hand-shaking) remains as failure for some reasons (e.g., poor WiFi connection), it blocks all subsequent calls. Small latency $\approx$ 10\unit{ms} in control commands makes a huge difference in the resultant performance. We thus modify the SDK to send direct control commands via serial communication.

In this paper, we only present the overview of the development. Details of the modifications and procedures are given in \texttt{http://goo.gl/lXRnU8}.

\subsection{Dynamic systems identification}
\label{sec:dynamics}
In this section, we present full dynamics system identification resulting from the simulator and experiments. We record input and output data; Virtual RC commands and attitude response while manual flight on an onboard computer. It is worth mentioning that time stamps should be synchronized and logging the actual control commands, not transmitter inputs which have the highest priority with the minimum latency.

\subsubsection{Input commands scaling}
\begin{figure}
\centering
\subfigure[]{\includegraphics[width=0.49\columnwidth]{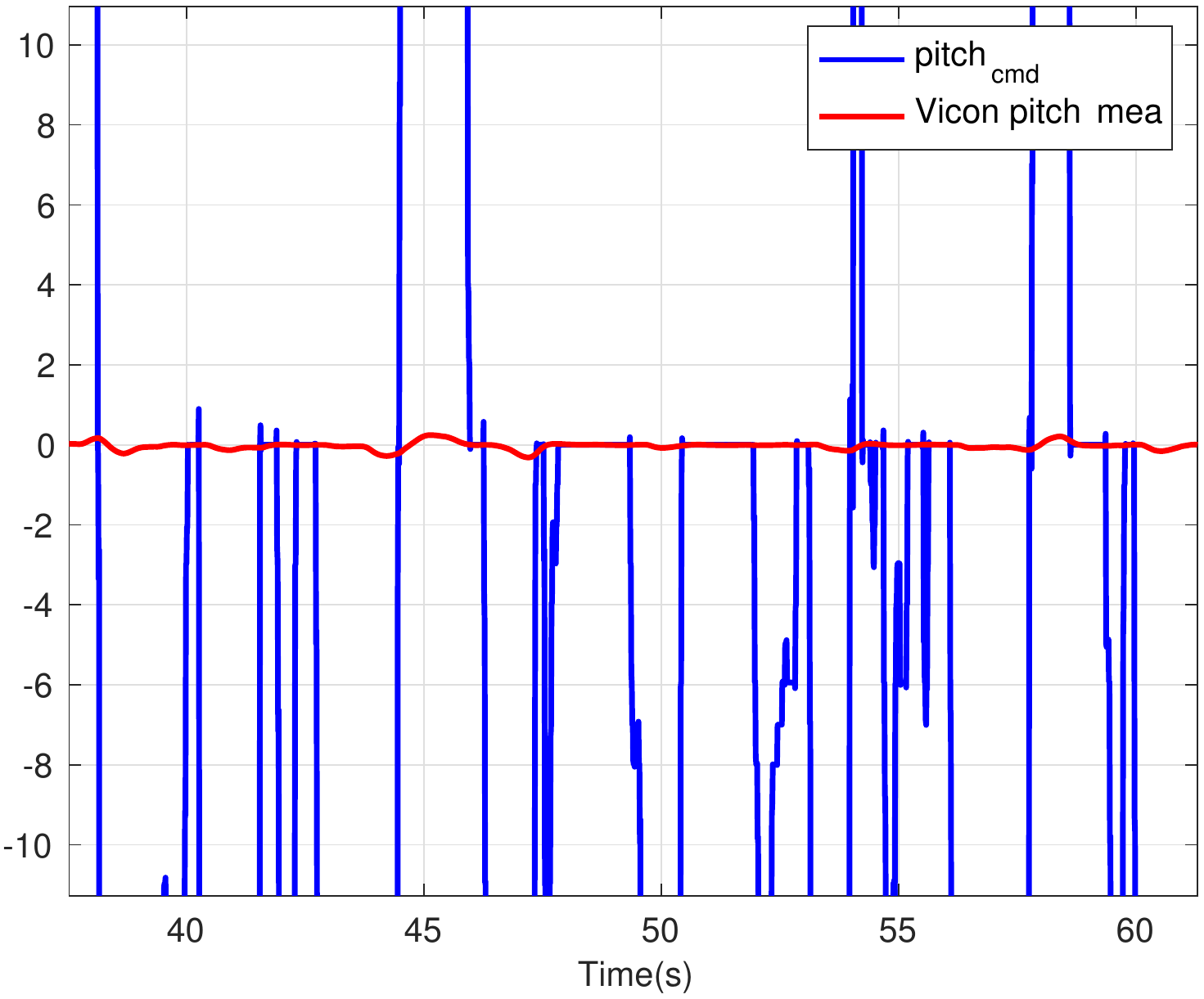}}
\subfigure[]{\includegraphics[width=0.49\columnwidth]{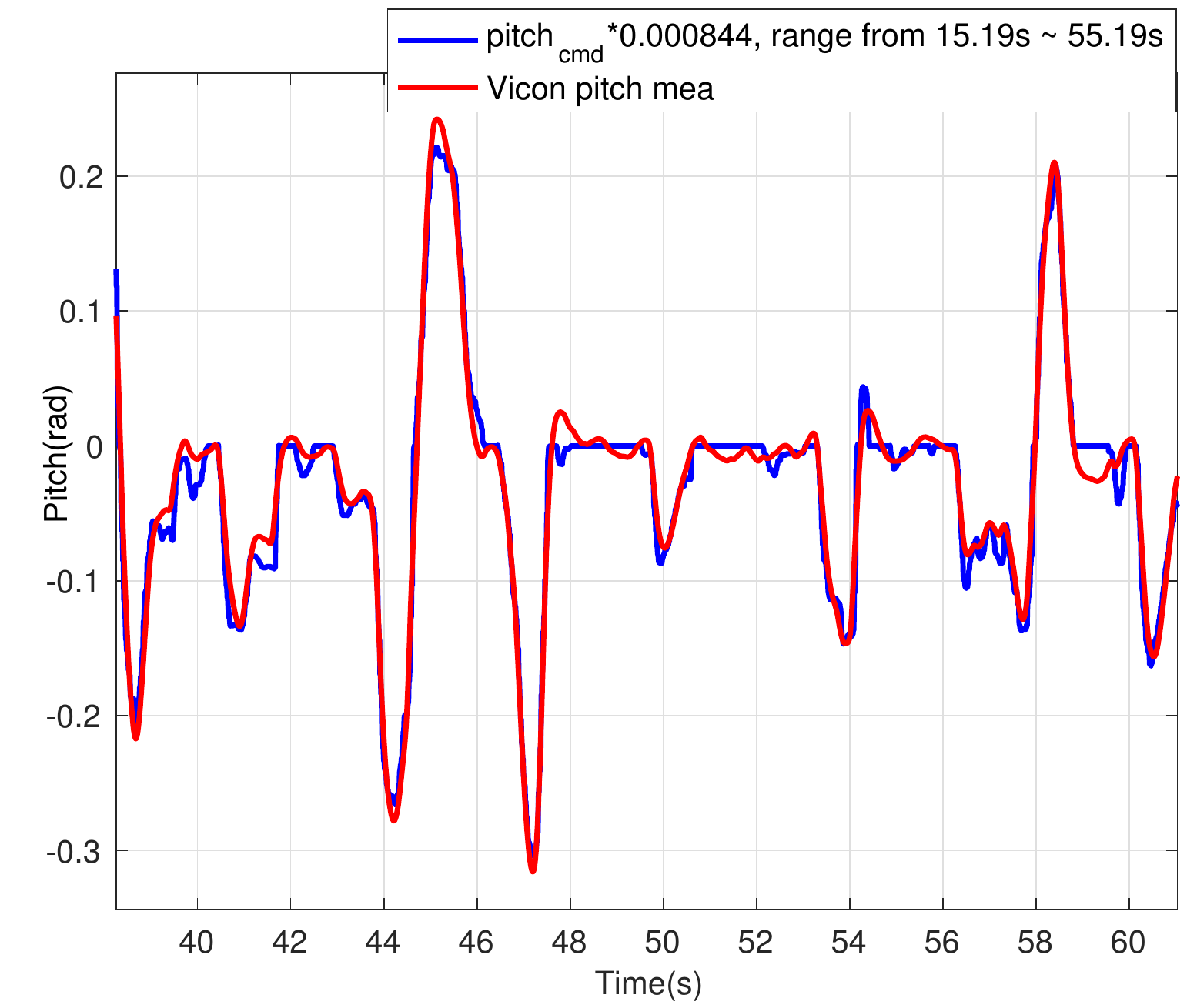}}
\centering
    \vspace{-5mm}
\caption{$u_{\mbox{\tiny $\phi$}}$ scaling parameter estimation. (a) and (b) are before and after the scaling.}
\label{fig:cmd_scaling}
    \vspace{-5mm}
\end{figure}

Prior to performing system identification, it is necessary to identify the relation between Virtual RC (actual control commands) and the corresponding attitude measurements from the IMU. This can be determined by linearly mapping with the maximum/minimum angles ($\pm30^\circ$), however there are small error in practice. This can be caused by variety of sources such as unbalanced platform, subtle differences in dynamics. We estimate these parameters using nonlinear least-squares optimization such that
\begin{align}
    \vspace{-2mm}
\boldsymbol{\lambda} ^{\star} := arg\underset{\boldsymbol{\lambda}}{min}\sum_{k=1}^{T}\|\boldsymbol{z}_{k}^{\mbox{\tiny Meas}}-\boldsymbol{\lambda} \boldsymbol{u}_{k}^{\mbox{\tiny cmd}}\|^{2}
\label{eq:scaleOptimisation}
    \vspace{-2mm}
\end{align}
where $T$ denotes the number of samples used for optimization. $\boldsymbol{\lambda}$ is 4$\times$1 vector containing roll, pitch, and yaw rate scaling parameters, $[\lambda_{\mbox{\tiny $\phi$}}, \lambda_{\mbox{\tiny $\theta$}}, \lambda_{\mbox{\tiny $\dot{\psi}$}}, \lambda_{\mbox{\tiny $\dot{\text{z}}$}}]^T$. $\boldsymbol{z}_{k}^{\mbox{\tiny Meas}}$ is 4$\times$1 of $[\phi, \theta, \dot{\psi}, \dot{\text{z}}]^T$ obtained from IMU and a motion capture device at 100\unit{Hz}. $\boldsymbol{u}_{k}^{\mbox{\tiny cmd}}$ is also 4$\times$1 vector of Virtual RC commands $[u_{\mbox{\tiny $\phi$}}, u_{\mbox{\tiny $\theta$}}, u_{\mbox{\tiny $\dot{\psi}$}}, u_{\mbox{\tiny $\dot{\text{z}}$}}]^T$. Note that it is difficult to measure vertical velocity using IMU, but there are three options; 1) motion capture device, 2) ultrasonic, barometer, and IMU fusion for vertical velocity estimation from the onboard flight controller, 3) using the simulator. The first approach is the most accurate but the third is useful and produces similar results as shown Table~\ref{tbl:scale} and Fig.~\ref{fig:cmd_scaling}. The input commands and output measurements are aligned with cross-correlation to remove delay between the two signals. This is acceptable since we estimate the signal magnitude. Fig.~\ref{fig:cmd_scaling} (a) shows original input command in blue and angle measurement in red. (b) displays results after scaling using estimated parameters from Table \ref{tbl:scale}. Linear scaling yields $\lambda_{\mbox{\tiny $\phi$}} = 7.93\times10^{-4}$ whereas the estimates are $\lambda_{\mbox{\tiny $\phi$}}=8.65\times10^{-4}$, and $\lambda_{\mbox{\tiny $\theta$}}=8.44\times10^{-4}$.
\begin{table}
\centering
\caption{Virtual RC scaling parameters}
    \vspace{-2mm}
\label{tbl:scale}
\begin{tabular}{ccc}
\textbf{Scale param} & \textbf{Experiment} & \textbf{Simulator} \\ \hline
$\lambda_{\mbox{\tiny $\phi$}}$                 & $8.65\times10^{-4}$          & $8.35\times10^{-4}$       \\ \hline
$\lambda_{\mbox{\tiny $\theta$}}$                & $8.44\times10^{-4}$            & $8.23\times10^{-4}$           \\ \hline
$\lambda_{\mbox{\tiny $\dot{\psi}$}} $            & $2.24\times10^{-3}$            & $3.23\times10^{-3}$           \\ \hline
$\lambda_{\mbox{\tiny $\dot{\text{z}}$}} $              & $2.65\times10^{-3}$      & $3.02\times10^{-3}$           \\ \hline
\end{tabular}
    \vspace{-5pt}
\end{table}

\subsubsection{Roll and pitch attitude dynamics}
Our linear MPC controller requires first order attitude dynamics for position control and the second order for the disturbances observer. We estimate the dynamics by recording the input and output at 100\unit{Hz} and logged two sets of dataset for model training and validation. 

We assume a low-level flight controller that can track the reference roll, $\phi^{*}$, and pitch, $\theta^{*}$, angles with first order behavior. The first order approximation provides sufficient information to the MPC to take into account the low-level controller behavior. We thus utilize classic system identification techniques such that:
\begin{align*}
		\frac{y(s)_{\mbox{\tiny $\phi$}}}{u(s)_{\mbox{\tiny $\phi$}}}=\frac{3.544}{s+2.118}, \;\;\;
		\frac{y(s)_{\mbox{\tiny $\theta$}}}{u(s)_{\mbox{\tiny $\theta$}}}=\frac{3.827}{s+2.43}
\end{align*}
$y(s)_{\mbox{\tiny $\phi$}}$ and $u(s)_{\mbox{\tiny $\phi$}}$ are IMU measurement and input commands in continuous-time space. The time constants for roll and pitch are, $\tau_{\mbox{\tiny $\phi$}}=0.472$, $\tau_{\mbox{\tiny $\theta$}}=0.472$ and DC gains are $k_{\mbox{\tiny $\phi$}}=1.673, k_{\mbox{\tiny $\theta$}}=1.575$. 

\begin{figure}
\centering
\subfigure[]{\includegraphics[width=0.49\columnwidth]{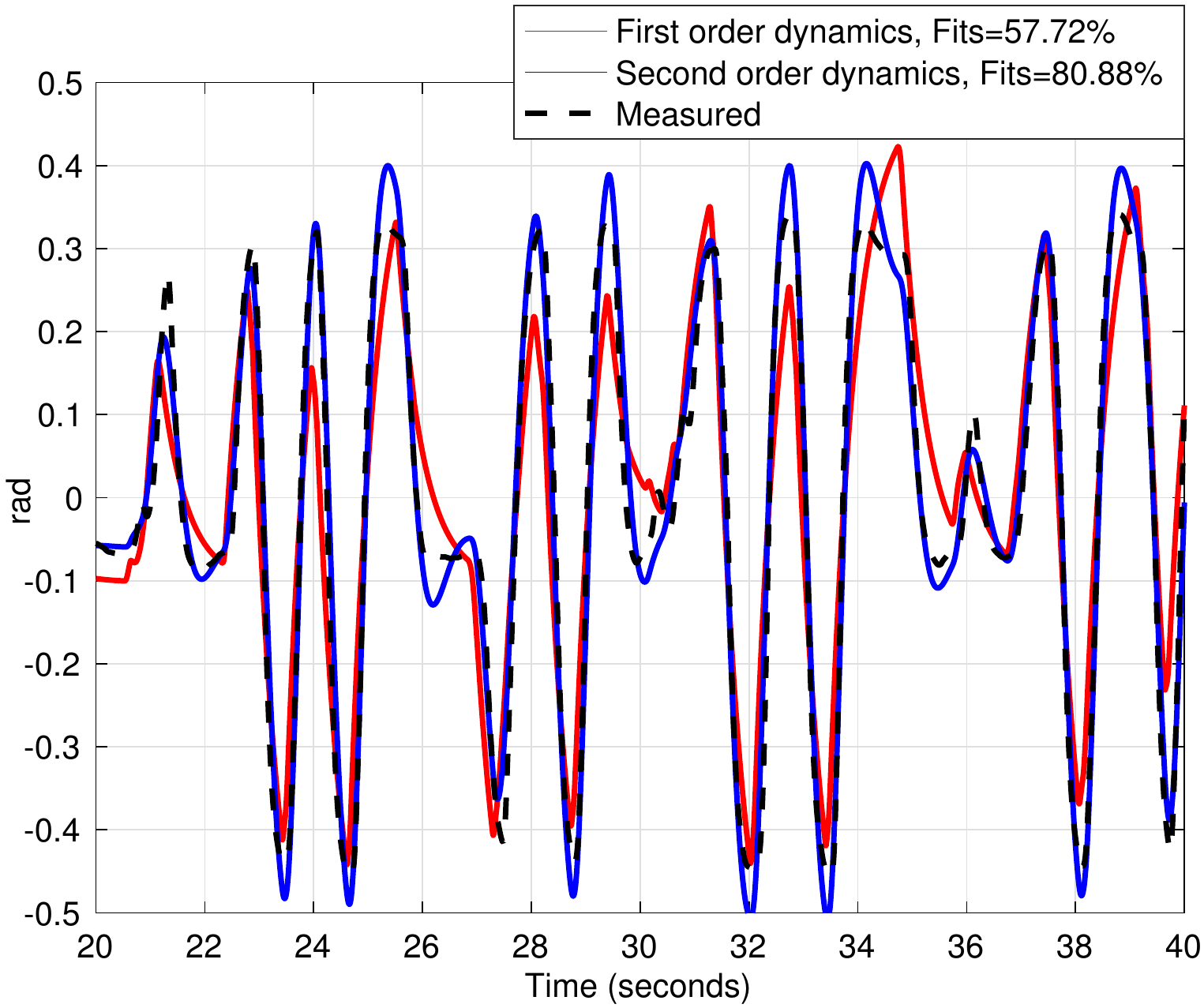}}
\subfigure[]{\includegraphics[width=0.49\columnwidth]{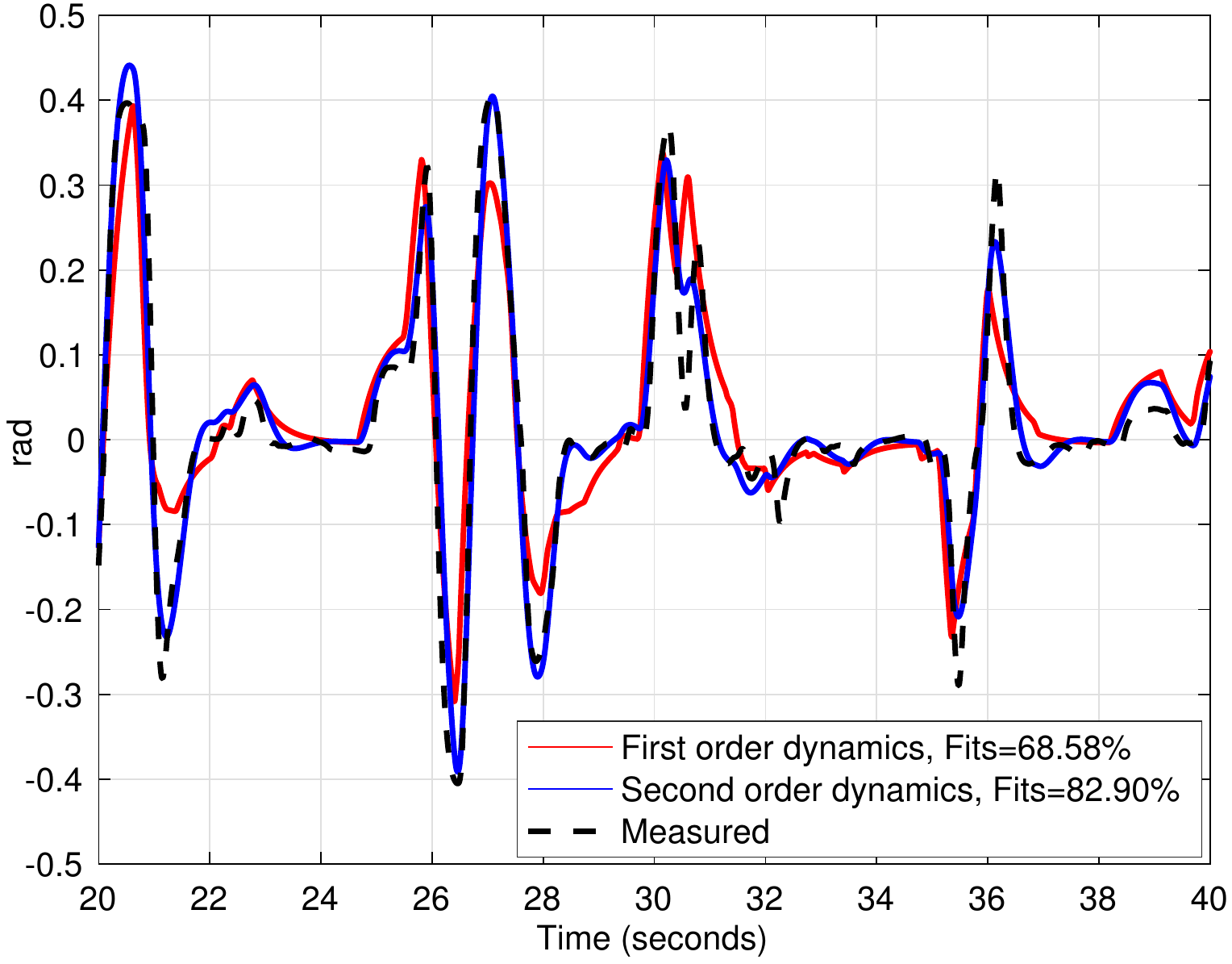}}
\centering
    \vspace{-5mm}
\caption{Measured and predicted pitch (a) and roll (b) angles for manual flight. The black denotes the measured angle, and the blue represents the model response. }
    \vspace{-5mm}
\label{fig:pitch_roll_dynamics}
\end{figure}

The identified first order dynamic models in continuous time space are discretized in MPC and will be addressed in the next section \ref{subsec:MPC}. We also performed second order dynamic system identification exploited by disturbances observer, and the dynamic models are 
\begin{align*}
		\frac{y(s)_{\mbox{\tiny $\phi$}}}{u(s)_{\mbox{\tiny $\phi$}}}=\frac{26.37}{s^2+5.32s+27.04},\;
		\frac{y(s)_{\mbox{\tiny $\theta$}}}{u(s)_{\mbox{\tiny $\theta$}}}=\frac{28.86}{s^2+6.00s+27.45}
\end{align*}
Their gain, $k_{\mbox{\tiny $\phi$}}$, damping, $\zeta_{\mbox{\tiny $\phi$}}$, and natural frequency, $\omega_{\mbox{\tiny $\phi$}}$ are presented in Table~\ref{tbl:dyna_sum}. Fig.~\ref{fig:pitch_roll_dynamics} shows measured attitude, estimated first and second order dynamics.

\setlength\tabcolsep{4pt}
\begin{table}
    \vspace{0mm}
\centering
\caption{M100 identified dynamics summary}
    \vspace{-2mm}
\label{tbl:dyna_sum}
\begin{tabular}{ccccc}
                           & $\phi$                               & $\theta$                            & $\dot{\psi}$ & $\dot{\text{z}}$ \\ \hline
\multirow{2}{*}{1st order} & $\tau_{\mbox{\tiny $\phi$}}=0.472$ & $\tau_{\mbox{\tiny $\theta$}}=0.472$ &   $\tau_{\mbox{\tiny $\dot{\psi}$}}=0.161$           &     $\tau_{\mbox{\tiny $\dot{\text{z}}$}}=0.334$               \\
                           & $k_{\mbox{\tiny $\phi$}}=1.673$    & $k_{\mbox{\tiny $\theta$}}=1.575$ &   $k_{\mbox{\tiny $\dot{\psi}$}}=1.057$           &       $k_{\mbox{\tiny $\dot{\stxt{z}}$}}=1.118$             \\ \hline
\multirow{3}{*}{2nd order} &  $k_{\mbox{\tiny $\phi$}}=0.975$                                  &   $k_{\mbox{\tiny $\theta$}}=1.052$                                &  $k_{\mbox{\tiny $\dot{\psi}$}}=1.079$            &    $k_{\mbox{\tiny $\dot{\text{z}}$}}=1.024$                \\
                           &     $\zeta_{\mbox{\tiny $\phi$}}=0.512$                               &     $\zeta_{\mbox{\tiny $\theta$}}=0.573$                              &    $\zeta_{\mbox{\tiny $\dot{\psi}$}}=1.898$          &    $\zeta_{\mbox{\tiny $\dot{\text{z}}$}}=0.718$                \\
                           &     $\omega_{\mbox{\tiny $\phi$}}=5.200$                               &   $\omega_{\mbox{\tiny $\theta$}}=5.239$                                &   $\omega_{\mbox{\tiny $\dot{\psi}$}}=23.448$           &      $\omega_{\mbox{\tiny $\dot{\text{z}}$}}=4.985$              \\ \hline
\end{tabular}
    \vspace{-5mm}
\end{table}

\subsubsection{Yaw and height dynamics}
\begin{figure}
\centering
\subfigure[]{\includegraphics[width=0.49\columnwidth]{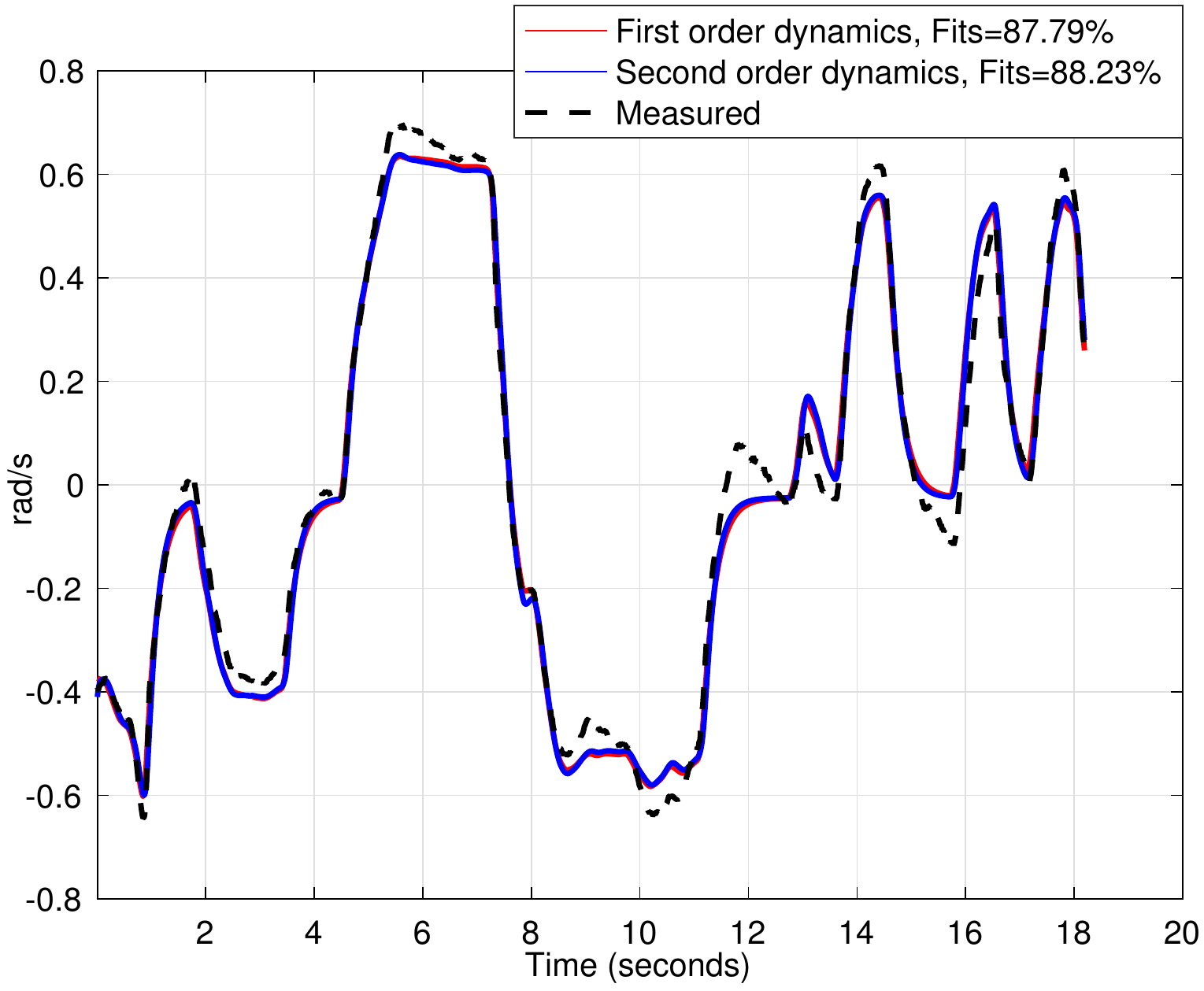}}
\subfigure[]{\includegraphics[width=0.49\columnwidth]{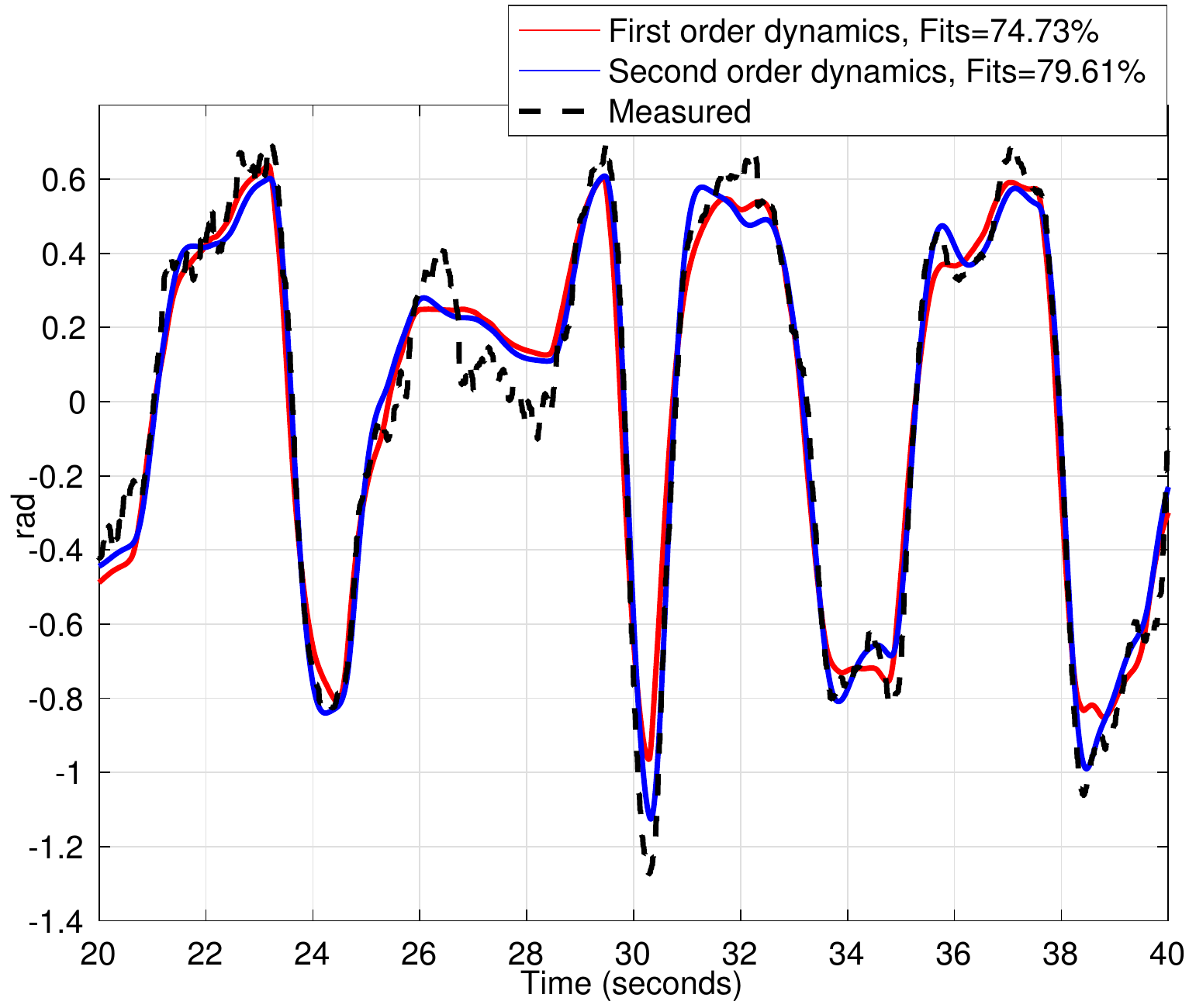}}
\centering
    \vspace{-5mm}
\caption{Measured and predicted yaw rate (a) and vertical velocity (b). The black denotes the measurement, and the blue represents the model response. }
\label{fig:yaw_height_dynamics}
    \vspace{-5mm}
\end{figure}

The input commands of yaw and height are rates, $u_{\mbox{\tiny $\dot{\psi}$}}, u_{\mbox{\tiny $\dot{\text{z}}$}}$, and the desired references are orientation, $\psi$ and position, z. This implies there are controllers that tracks the desired yaw rate, $\dot{\psi}^{*}$, and height velocity, $\dot{\text{z}}^{*}$. Their first order dynamics are
\begin{align*}
		\frac{y(s)_{\mbox{\tiny $\dot{\psi}$}}}{u(s)_{\mbox{\tiny $\dot{\psi}$}}}=\frac{5.642}{s + 5.268},\;
		\frac{y(s)_{\mbox{\tiny $\dot{\text{z}}$}}}{u(s)_{\mbox{\tiny $\dot{\text{z}}$}}}=\frac{3.342}{s + 2.99}
\end{align*}
Similarly, the second order dynamics for $\dot{\psi}$ and $\dot{\text{z}}$ are identified as
\begin{align*}
        \frac{y(s)_{\mbox{\tiny $\dot{\psi}$}}}{u(s)_{\mbox{\tiny $\dot{\psi}$}}}=\frac{593.3}{s^2 + 89.0s + 549},\;
        \frac{y(s)_{\mbox{\tiny $\dot{\text{z}}$}}}{u(s)_{\mbox{\tiny $\dot{\text{z}}$}}}=\frac{25.43}{s^2 + 7.16s + 24.8}
\end{align*}
$y(s)_{\mbox{\tiny $\dot{\psi}$}}$ is obtained from the built-in IMU, gyro measurement along z-axis, and $y(s)_{\mbox{\tiny $\dot{\text{z}}$}}$ is provided by a motion capture device. It is also feasible to identify the height dynamics by utilizing vertical velocity estimation from N1 flight controller. Fig.~\ref{fig:yaw_height_dynamics} shows the identified first and second order dynamics for yaw rate (a) and vertical velocity (b). The models fit close to output (dotted line) and they are utilized by controllers presented in the next subsection.

\subsection{Control}
\subsubsection{Linear-MPC for horizontal x and y}\label{subsec:MPC} 
The control of the lateral motion of the vehicle is based on a Linear Model Predictive Control (MPC) \cite{kamelmpc2016}. The vehicle dynamics are linearized around the hovering condition. We define the state vector to be $ \boldsymbol{x} = \left(\text{x}, \text{y}, v_{\stxt{x}}, v_{\stxt{y}}, {}^{\mathcal{W}}\phi, {}^{\mathcal{W}}\theta \right)$ and the control input vector to be $\boldsymbol{u} = \left({}^{\mathcal{W}}u_{\mbox{\tiny $\phi$}}, {}^{\mathcal{W}}u_{\mbox{\tiny $\theta$}} \right)$. We also define the reference state sequence as $ X^{T}_{\stxt{ref}} = \left[ \boldsymbol{x}^{T}_{\stxt{ref:0}}, \dots,  \boldsymbol{x}^{T}_{\stxt{ref:N}} \right]^{T} $, the control input sequence as $ U = \left[ \boldsymbol{u}^{T}_{\stxt{0}}, \dots ,\boldsymbol{u}^{T}_{\stxt{N-1}} \right]^{T} $, and the steady state input sequence $ U_{\stxt{ref}} = \left[ \boldsymbol{u}^{T}_{\stxt{ref:0}}, \dots, \boldsymbol{u}^{T}_{\stxt{ref:N-1}} \right]^{T} $. $\bs{x}_{\stxt{ref:k}}, \bs{u}_{\stxt{ref:k}}$ are the $\text{k}_{\stxt{th}}$ reference state and control input. Every time step, the following optimization problem is solved:
\begin{equation} \label{eq:mav_linear_mpc_opt}
\begin{aligned}
\min_{U,X} & 
\sum_{k=0}^{N-1} \left( \left(\bs{x}_{\stxt{k}} - \bs{x}_{\stxt{ref:k}}\right)^{T}Q_{\stxt{x}}\left(\bs{x}_{\stxt{k}} - \bs{x}_{\stxt{ref:k}}\right) \right.\\ 
+ & \left(\bs{u}_{\stxt{k}} - \bs{u}_{\stxt{ref:k}}\right)^{T}R_{\stxt{u}}\left(\bs{u}_{\stxt{k}} - \bs{u}_{\stxt{ref:k}}\right) \\+ & \left(\bs{u}_{\stxt{k}} - \bs{u}_{\stxt{k-1}}\right)^{T}R_{\Delta}\left(\bs{u}_{\stxt{k}} - \bs{u}_{\stxt{k-1}}\right)\left. \right)  \\  + & \left(\bs{x}_{\stxt{N}} - \bs{x}_{\stxt{ref:N}}\right)^{T}P\left(\bs{x}_{\stxt{N}} - \bs{x}_{\stxt{ref:N}}\right)\\
&\begin{aligned}
\text{subject to} &
& & \bs{x}_{\stxt{k+1}} = A\bs{x}_{\stxt{k}} + B\bs{u}_{\stxt{k}} + B_{\stxt{d}}{d}_{\stxt{k}};\\
& & & d_{\stxt{k+1}} = d_{\stxt{k}}, \quad \text{k}=0, \dots, N-1\\
& & & \bs{u}_{\stxt{k}} \in \mathbb{U} \\
& & & \bs{x}_{\stxt{0}} = \bs{x}\left( {t_{\stxt{0}}}\right), \quad d_{\stxt{0}} = d\left( t_{\stxt{0}} \right) .
\end{aligned}
\end{aligned}
\end{equation}

\noindent where $ Q_{\stxt{x}} \succeq 0 $ is the penalty on the state error, $ R_{\stxt{u}} \succ 0 $ is the penalty on control input error,   $ R_{\Delta} \succeq 0 $ is a penalty on the control change rate and $ P $ is the terminal state error penalty. The $\succeq$ operator denotes positive definiteness of a matrix\footnote{https://en.wikipedia.org/wiki/Positive-definite\_matrix}. $d_{\stxt{k}}$ is the estimated external disturbances. Note that the attitude angles $\phi, \theta$ are rotated into the inertial frame to get rid of the vehicle heading $\psi$.

A high performance solver has been generated to solve the optimization problem \eqref{eq:mav_linear_mpc_opt} using the FORCES \cite{FORCESPro} framework. The solver is running in real-time for a prediction horizon of $N=20$ steps. Moreover, to achieve an offset-free tracking, the external disturbances $d_{k}$ has to be estimated and provided to the controller each time step. These disturbances include external forces (the wind for instance) and also a modeling error. The disturbances are estimated using an augmented Extended Kalman Filter (EKF) based on the second order dynamics model.

\subsubsection{Vertical (altitude) and yaw control}

We use a standard PID controller to compute velocity command as
\begin{equation}\label{height_pid} 	
        u_{\mbox{\tiny $\dot{\text{z}}$}}(t)=K_{\stxt{p}}e_{\stxt{z}}(t)+K_{\stxt{i}}\int_{0}^{t}e_{\stxt{z}}(\tau)d\tau+K_{\stxt{d}}\frac{d}{dt}e_{\stxt{z}}(t)
\end{equation}
where $e_{\stxt{z}}(t)=\text{z}^*-\text{z}$, is position error, $\text{z}^*$ denotes the desired height, and $\text{z}$ is the current measurement. Given the height dynamics model, we simulate a PID controller and tune parameters ($K_{\stxt{p}}=1.963, K_{\stxt{i}}=6.190, K_{\stxt{d}}=0.156, I_{\stxt{min}}=-0.2 \text{, and } I_{\stxt{max}}=0.2$). As mentioned in the previous section \ref{sec:deadZone}, there exists the height dead zone where all control commands are disregarded and the height dead zone is wider than other control inputs. The integral term helps to compensate the steady-state error.
Yaw control is simply achieved by using P controller as
\begin{equation}\label{height_pid}
u_{\mbox{\tiny $\dot{\psi}$}}(t)=K_{\mbox{\tiny $\psi$}}e_{\mbox{\tiny $\psi$}}(t)
\end{equation}
where $e_{\mbox{\tiny $\psi$}}(t)=\psi^*-\psi$, is heading angle error and $K_{\mbox{\tiny $\psi$}}$ is a proportional gain.

\section{Experimental results}\label{sec:results}
In this section, we present implementation details; hardware and software setup and control performance evaluation for simulation and actual experiments.
\subsection{Hardware Setup}
The DJI Matrice 100 quadcopter carries an Intel NUC 5i7RYH (i7-5557U, 3.1\unit{GHz} dual cores, 16\unit{GB} RAM), running Ubuntu Linux 14.04 and ROS Indigo onboard \cite{Quigley:2009aa} as shown in Fig.~\ref{fig:M100}.
A Vicon motion capture system consisting of 6 IR cameras provides 6 DOF pose of the quadcopter target at 100 \unit{Hz}, used for control. The quadcopter is also equipped with a flight controller, N1, embedded an onboard IMU which provides vehicle orientation, acceleration, and angular velocity at 100 \unit{Hz} to the computer via 921,600\unit{bps} USB-to-serial communication. Control commands are also transmitted at 100\unit{Hz} through the serial bus.

The total system mass is 3.3\unit{kg} and WiFi is used for communicating with the quadcopter using a ground control laptop via ROS and a customized DJI-SDK ROS interface. The ground station, Vicon server, and onboard computer are time synchronized via chrony. The vehicle carries 0.92\unit{kg} payload with the onboard computer, a guidance system, and a gimbal camera. 6 cells LiPo battery, 22.2\unit{V}, 4500\unit{mAh}  powers the vehicle and the total flight time is around 14\unit{mins} with small angle of attack $\approx\pm20^{\circ}$. 

\subsection{Software Setup}

\begin{figure}
\begin{center}
    \includegraphics[width=0.8\columnwidth]{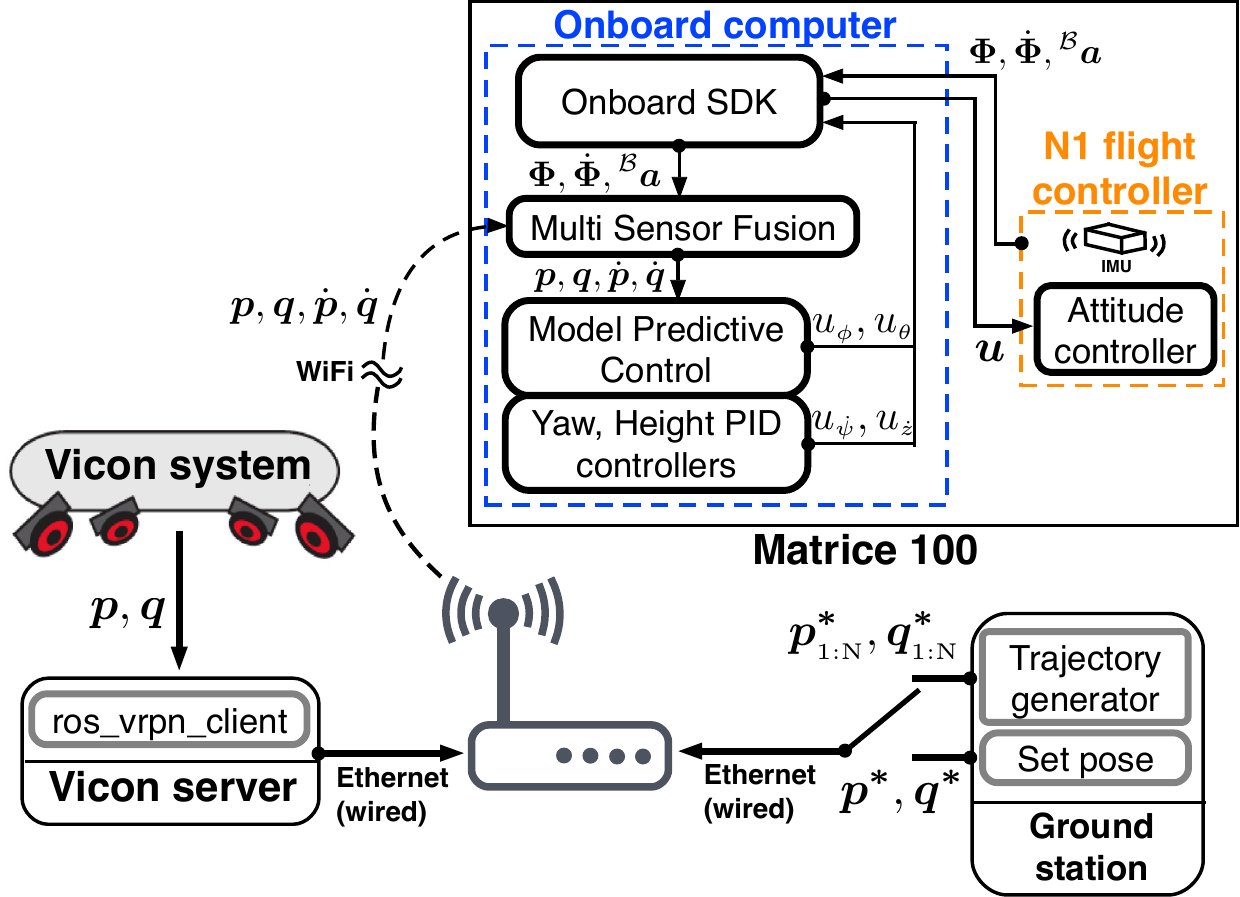}
\end{center}
    \vspace{-5mm}
\caption{Software packages implementation using ROS.}
\label{fig:diagram}
    \vspace{-2mm}
\end{figure}

We integrate the system using ROS as shown in Fig.~\ref{fig:diagram}. Each box represents a ROS node and runs at 100\unit{Hz}. The Vicon server publishes position, orientation, translational and angular velocity as denoted $[\boldsymbol{p},\boldsymbol{q},\dot{\boldsymbol{p}},\dot{\boldsymbol{q}}]$ using ros\_vrpn\_client.
The data is subscribed by the Multi-Sensor Fusion (MSF) framework\cite{Weiss:2011aa} to filter noisy measurement and to compensate for possible delay in the WiFi network connection. The ground station sets either a goal position as denoted $[\boldsymbol{p^{*}},\boldsymbol{q^{*}}]$ for position and orientation or N sequences, $[\boldsymbol{p_{\mbox{\tiny 1:N}}^{*}},\boldsymbol{q_{\mbox{\tiny 1:N}}^{*}}]$, generated by the trajectory generator\cite{burri2016maximum}.

The SDK runs on the onboard computer and receives IMU measurements, $[\boldsymbol{\Phi},\dot{\boldsymbol{\Phi}},{}^{\mathcal{B}}\boldsymbol{a}]$, orientation, angular velocity, acceleration in $\mathcal{B}$ coordinate from the N1 flight controller. It also sends the calculated control commands to the attitude controller.

\subsection{Experiments Setup}
For control performance evaluation, we conduct both simulation and real world experiments. For the simulation experiments, we use a hardware-in-loop simulator provided the manufacturer for and trajectory following task has been conducted. For the real world experiments, we perform 3 tasks; hovering, step response, and trajectory following. To demonstrate, the robustness of the controllers, we generate wind disturbances using a fan that has 260\unit{W} and 300\unit{m^3/min} air flow. This produces 11-11.5\unit{m/s} disturbance at the hovering position measured by an anemometer. Each task has two results, i.e., with/without wind disturbances.

We use root-mean-square (RMS) error metric between the reference and actual position and orientation measured by a motion capture device for performance evaluation. Euclidian distance is used for 3D pose RMS error calculation. Table~\ref{tbl:results_summary} summarizes simulation and experimental results.

\subsection{Control performance evaluation}
In this subsection, we present quantitative control performance evaluation using RMS error while performing 3 tasks, i.e., hovering, step response, and trajectory following, and qualitative results for step response and trajectory following.

\setlength\tabcolsep{2pt}

\begin{table}[]
\centering
\caption{Control performance summary}
    \vspace{-2mm}
\label{tbl:results_summary}
\begin{tabular}{ccccccccc}
\hline
& \multicolumn{2}{c}{\textbf{Hovering}} & \multicolumn{2}{c}{\textbf{Step response}} & \multicolumn{2}{c}{\textbf{Trj. following}} &(Simul) &Unit      \\ \hline
Pose                   &                              \textbf{0.039} &    0.041                           & 0.394                    &     0.266                 &    0.080                    &       0.103     &   0.108          & \unit{m}    \\ \hline
x                        &                              \textbf{0.021} &         0.027                      &  0.259                   &      0.127                &       0.058                 &        0.066      &   0.047           & \unit{m}    \\ \hline
y                        &                    0.016           &          \textbf{0.015}                     &   0.295                  &      0.226                &      0.043                  &        0.059      &   0.053           & \unit{m}    \\ \hline
z                        &                      0.029         &          \textbf{0.026}                     &   0.034                  &     0.059                 &     0.035                   &        0.053      &   0.082           & \unit{m}    \\ \hline
roll                     &                       0.392        &            1.044                   &   ---                  &         ---             &         ---               &         &   ---                & \unit{deg}  \\ \hline
pitch                    &                      0.618         &          0.697                     &  ---                   &       ---               &        ---                &          &   ---              & \unit{deg}  \\ \hline
yaw                      &                   \textbf{1.087}            &          1.844                     &  1.141                   &     2.165                 &        1.539                &      2.876         &   26.267 & \unit{deg}  \\ \hline
Duration                 &           15-75                    &        15-75                      &   20-120                  &   20-120                   &   30-80                     &    20-70          &   5-100           & \unit{s}    \\ \hline
Wind               & ---                           &         11-11.5        & ---                 &       11-11.5               & ---                    &     11-11.5          &   ---          & \unit{m/s}  \\ \hline
\end{tabular}
\vspace{-5mm}
\end{table}

\subsubsection{Simulation results for trajectory following}
\begin{figure}
\centering
\subfigure[]{\includegraphics[width=0.49\columnwidth]{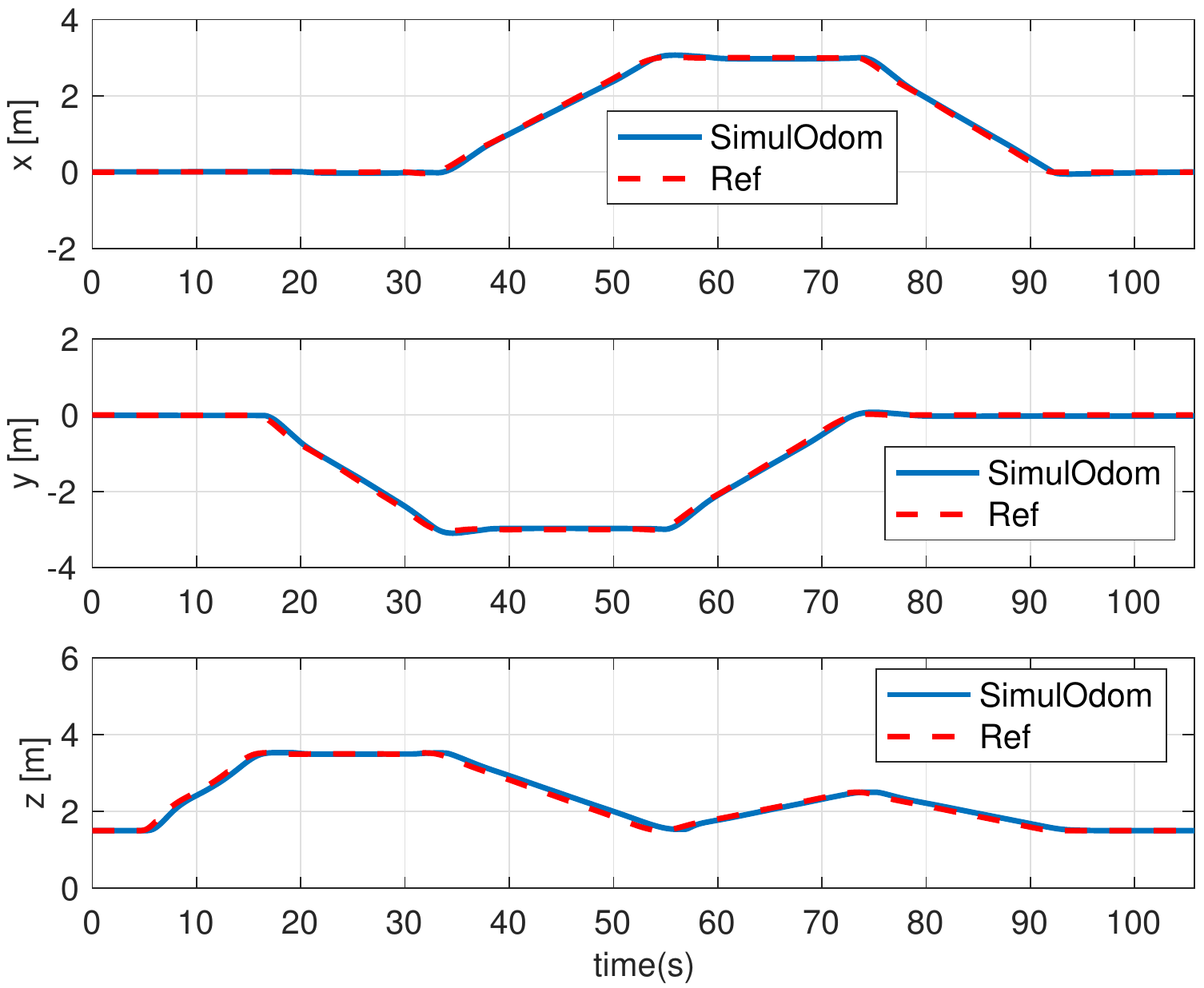}}
\subfigure[]{\includegraphics[width=0.49\columnwidth]{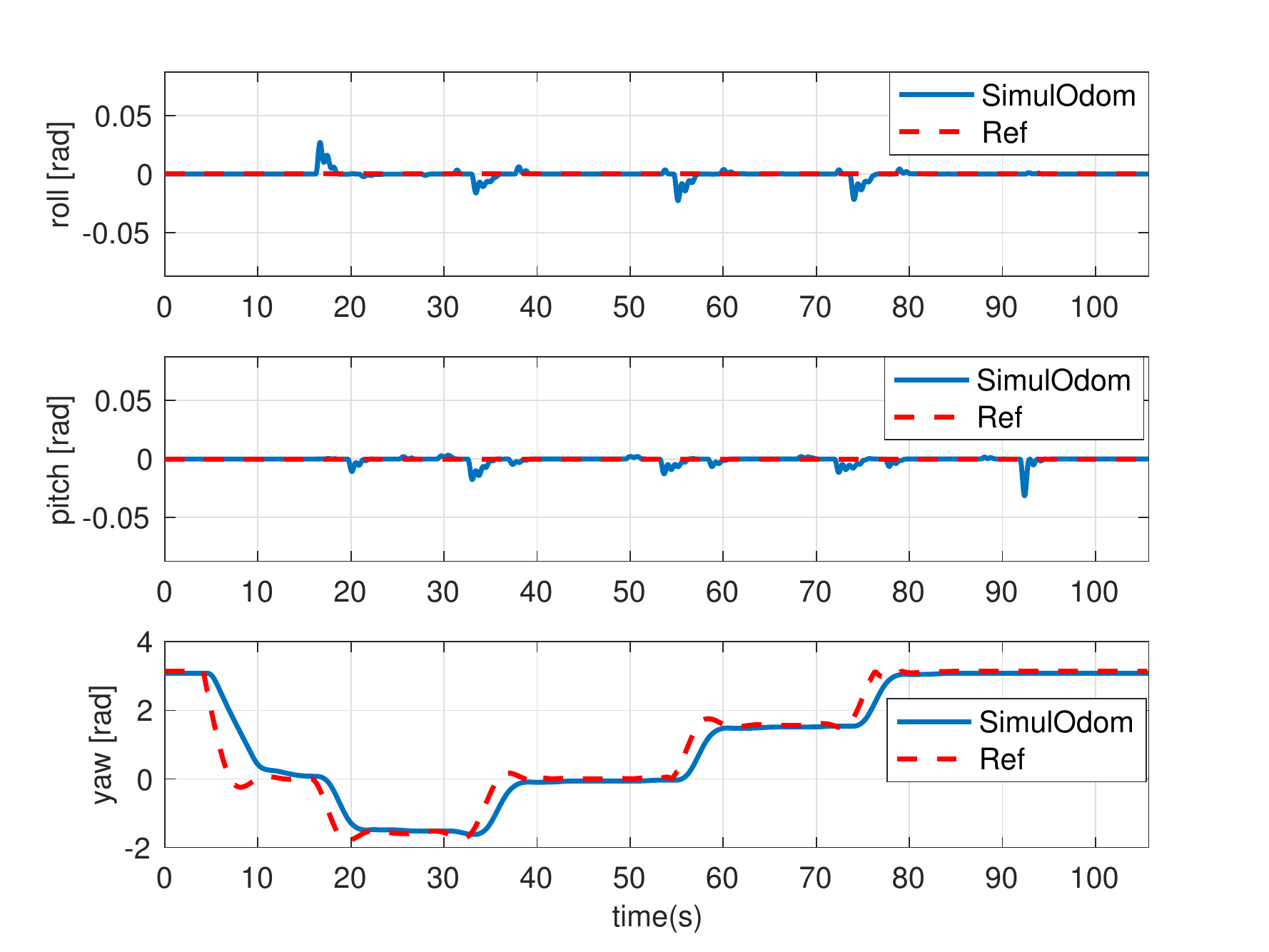}}
\centering
\vspace{-3mm}
\caption{Trajectory points following control performance with the simulator.}
\label{fig:simul_wp}
\vspace{-5mm}
\end{figure}

Before actual experiments, we develop a system shown in Fig.~\ref{fig:diagram} using a simulator. This provides noise-free simulated position, orientation, translational, angular velocity, and acceleration at 100\unit{Hz}. We also conduct system identification over the simulator and it turns out that dynamics models are close to the real system. This implies we can make use of the identified models as a good initial guess for controller parameter tuning. Fig.~\ref{fig:simul_wp} (a) and (b) show position and orientation control performance while performing trajectory following. 4 goal positions and orientations are set to excite horizontal, vertical, and heading motions. The red denotes reference and blue is the vehicle trajectory. Note that yaw has a different scale to pitch and roll due to large variation. We omit roll and pitch in RMS calculation because the reference horizontally moves without rotation. VTOL MAV platforms such as quad-, hexa-rotor cannot translate without tilting toward the direction of travel. The results are given in Table~\ref{tbl:results_summary}.

\subsubsection{Experiments results for hovering}
\begin{figure}
\centering
\subfigure[]{\includegraphics[width=0.49\columnwidth]{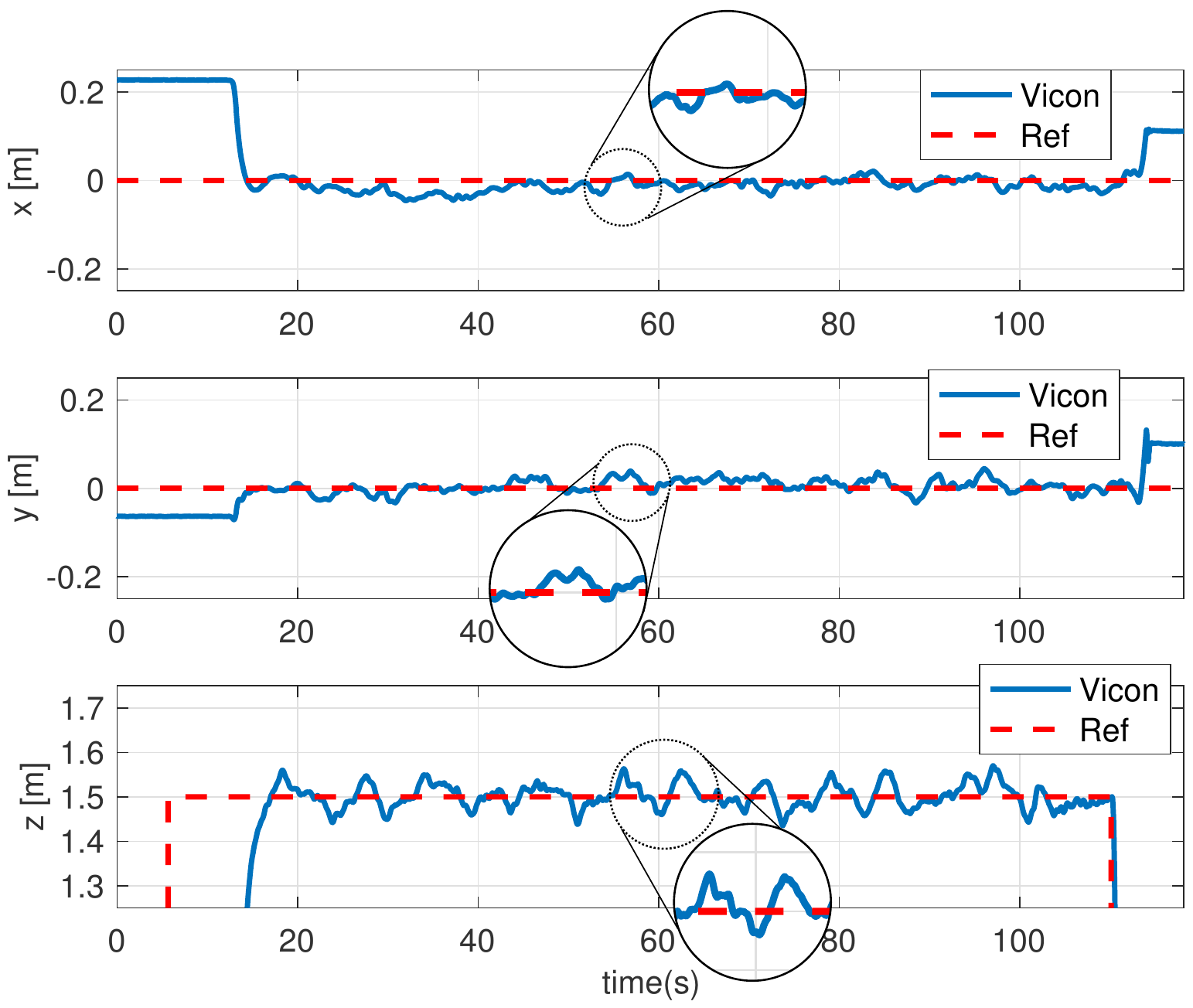}}
\subfigure[]{\includegraphics[width=0.49\columnwidth]{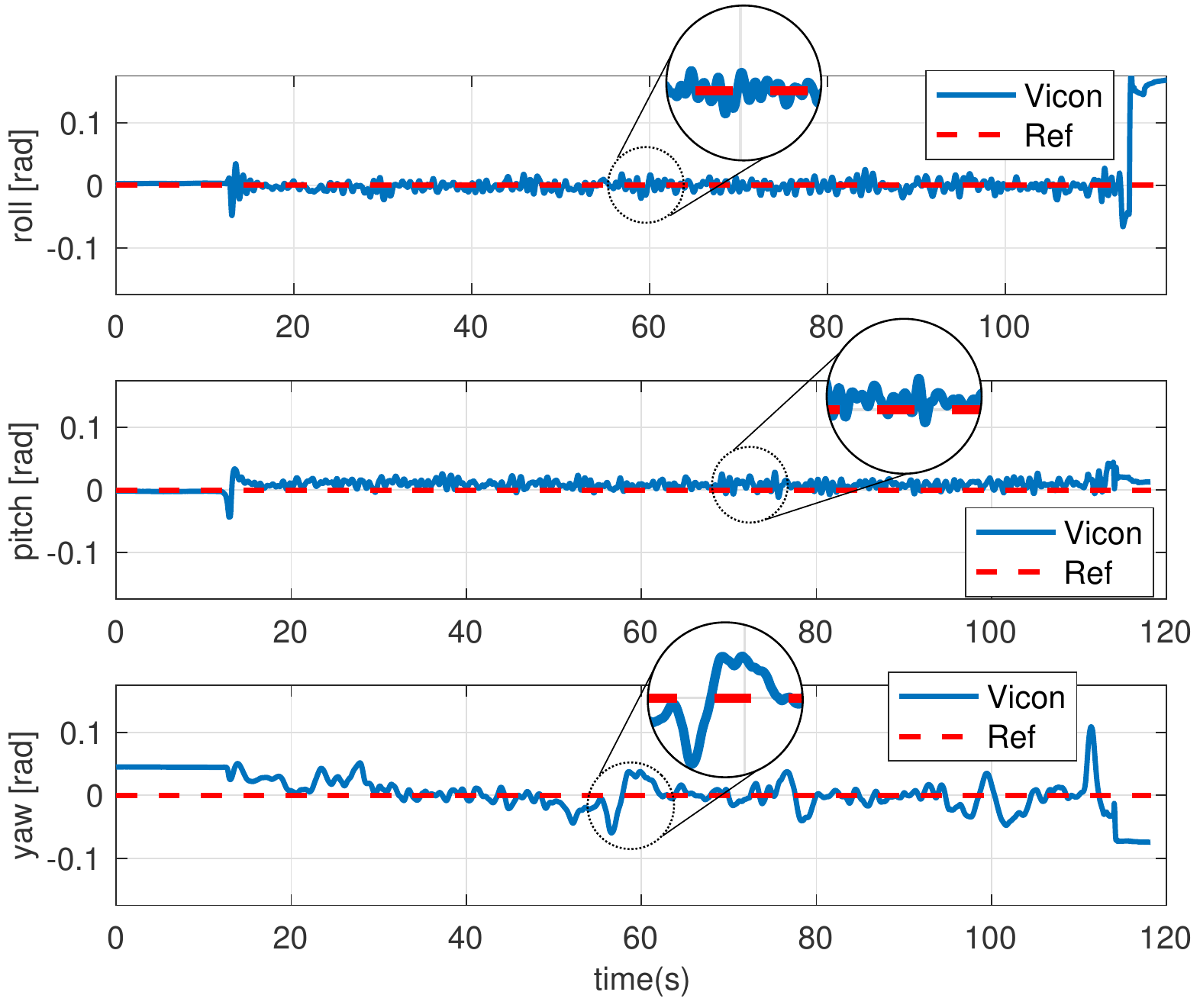}}
\subfigure[]{\includegraphics[width=0.49\columnwidth]{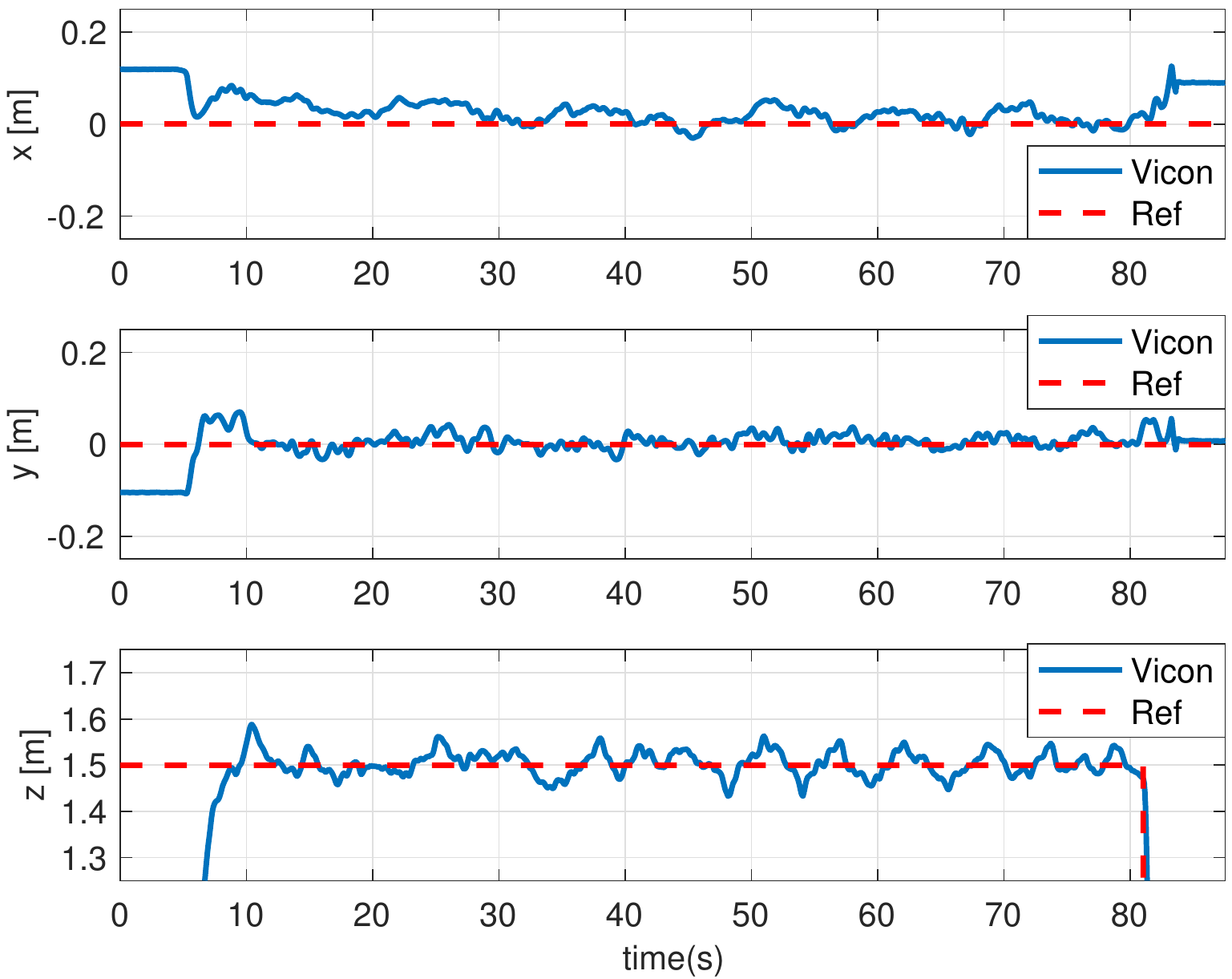}}
\subfigure[]{\includegraphics[width=0.49\columnwidth]{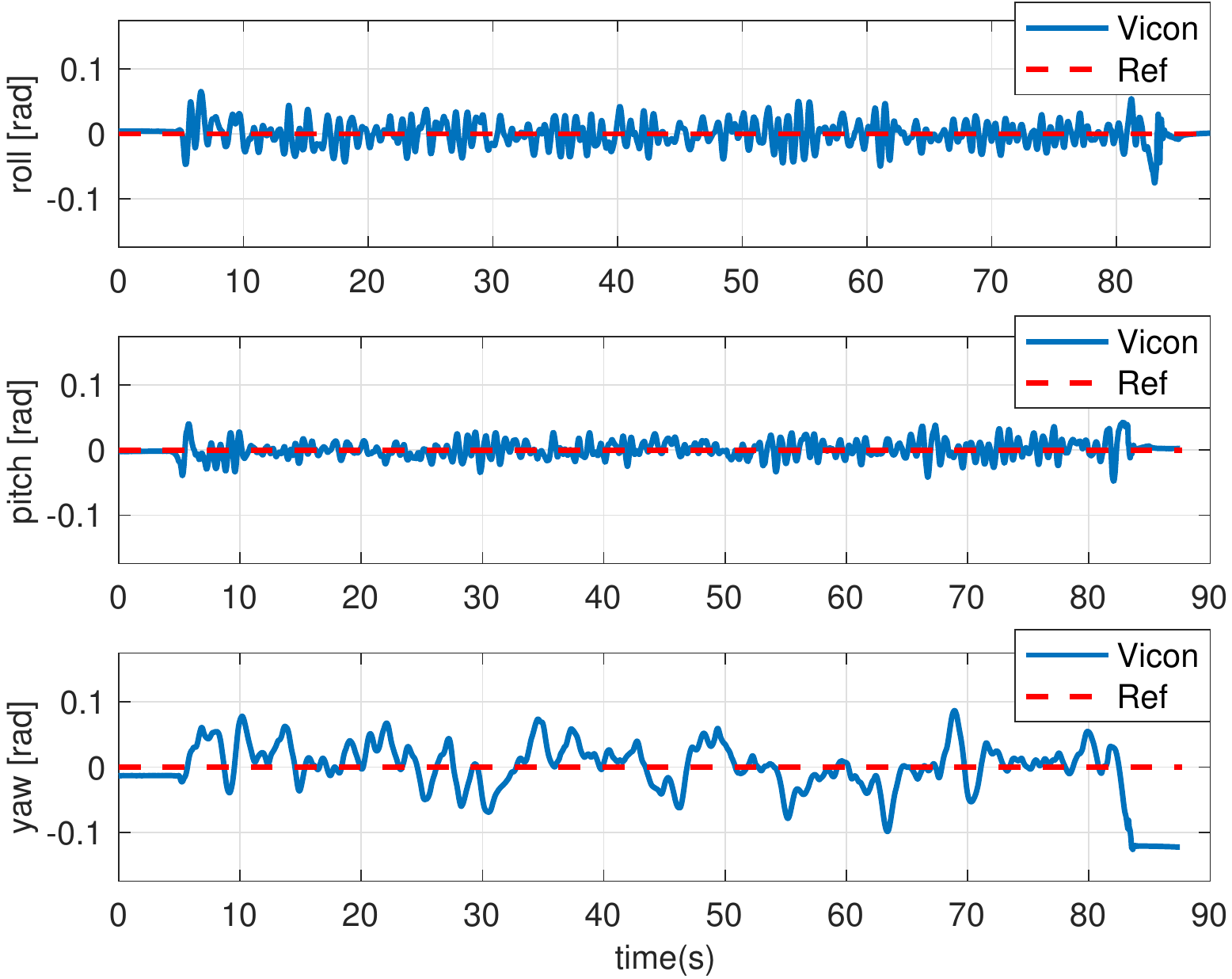}}
\centering
\vspace{-5mm}
\caption{Hovering control performance without/with wind disturbances.}
\label{fig:hovering}
\vspace{-3mm}
\end{figure}

Fig.~\ref{fig:hovering} shows hovering results without any wind disturbances (a), (b) for position and orientation respectively, and with disturbances (c), and (d). Noticeable areas are magnified due to the small scale of plots. We can clearly see the presence of wind disturbances affect to control performance. Especially, the variation in yaw and attitude are significant since a fan is located at South-East of the vehicle. As shown in Table~\ref{tbl:results_summary}, we achieve competitive results while hovering. Interestingly, the position errors for x, y, and z are consistent even with wind disturbances, 11-11.5\unit{m/s}. The force acting on the platform in the wind is too small to push the 3.3\unit{kg} flying platform.

\subsubsection{Experiments results for step response}
\begin{figure}
\centering
\subfigure[]{\includegraphics[width=0.49\columnwidth]{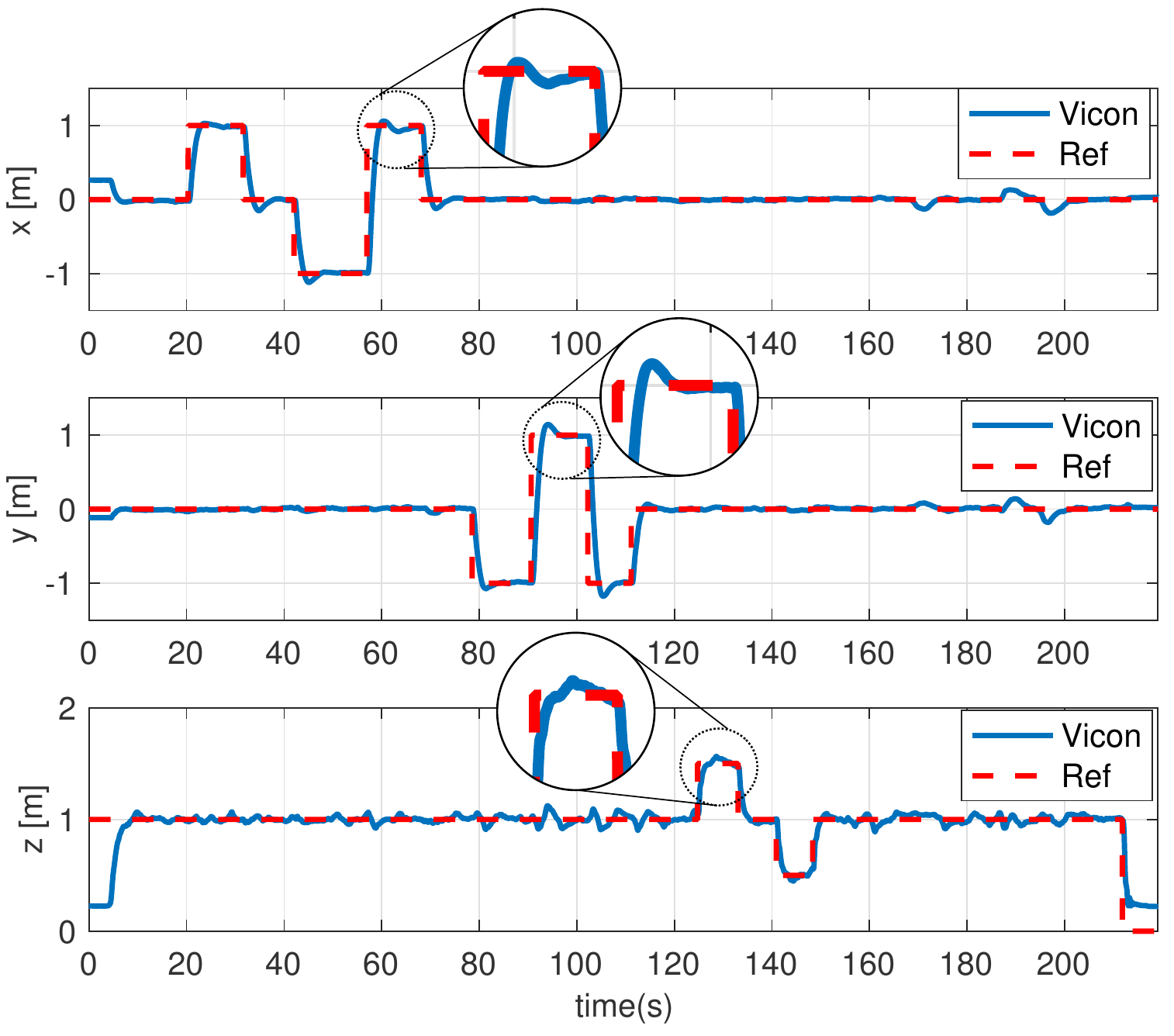}}
\subfigure[]{\includegraphics[width=0.49\columnwidth]{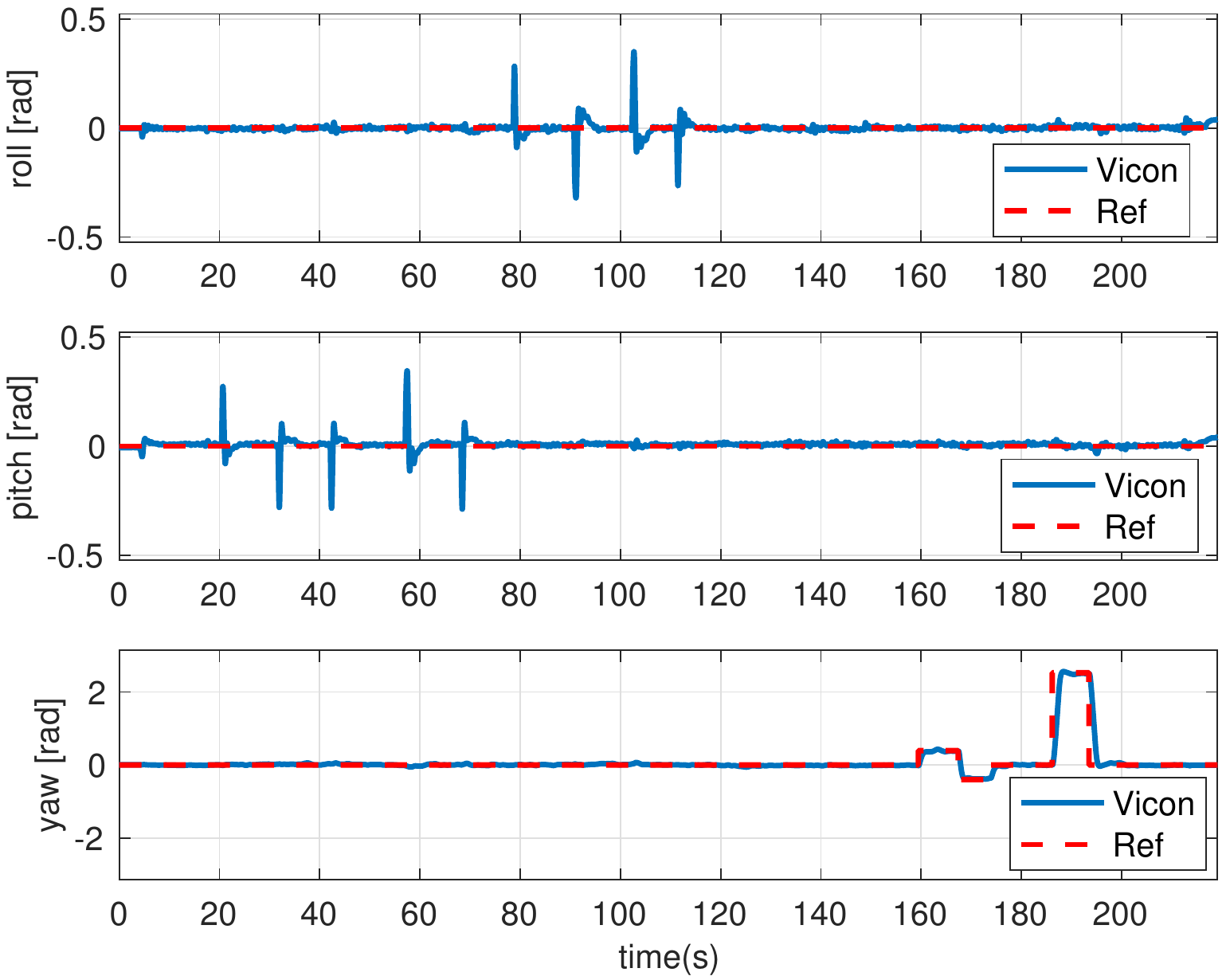}}
\subfigure[]{\includegraphics[width=0.49\columnwidth]{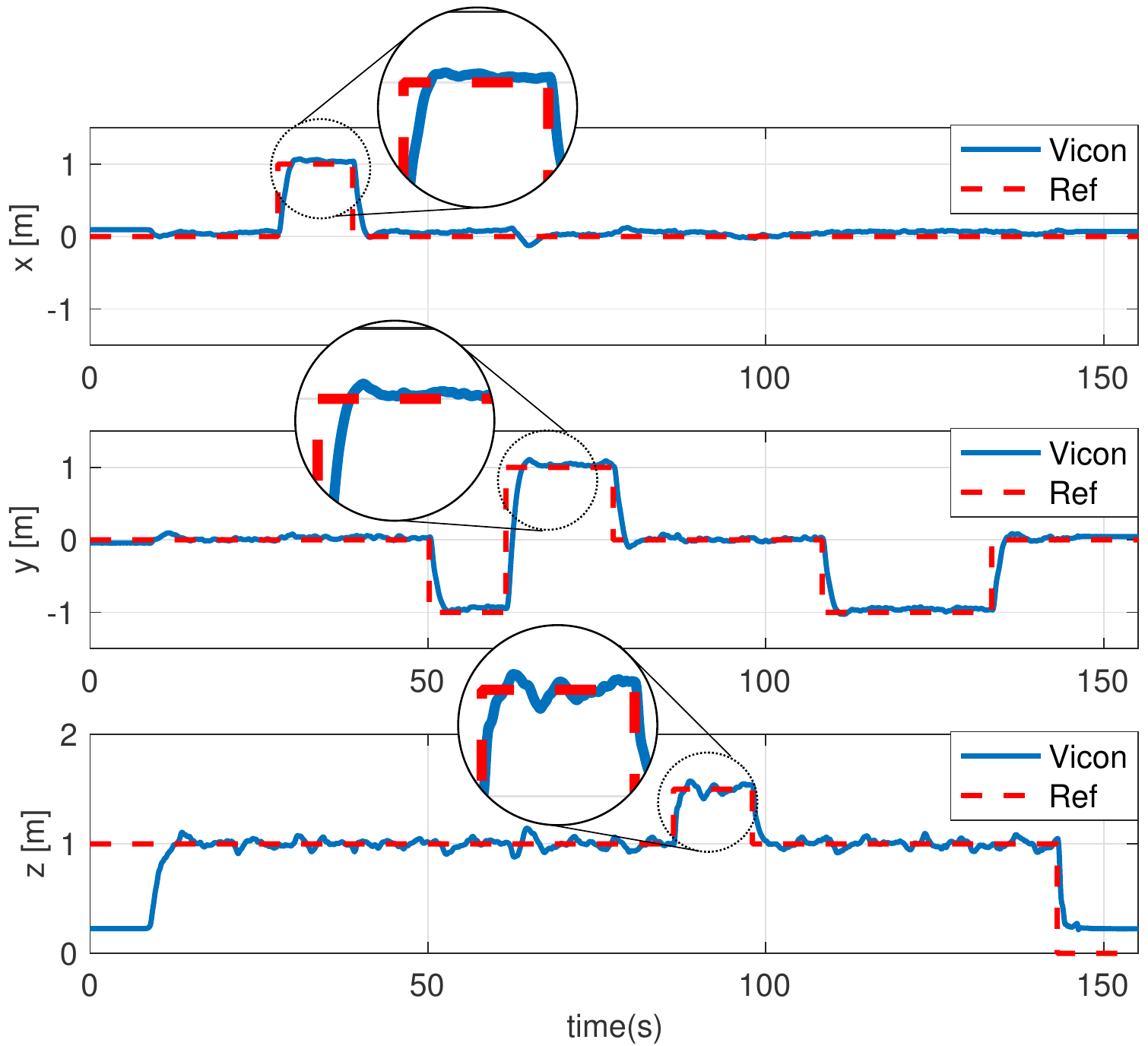}}
\subfigure[]{\includegraphics[width=0.49\columnwidth]{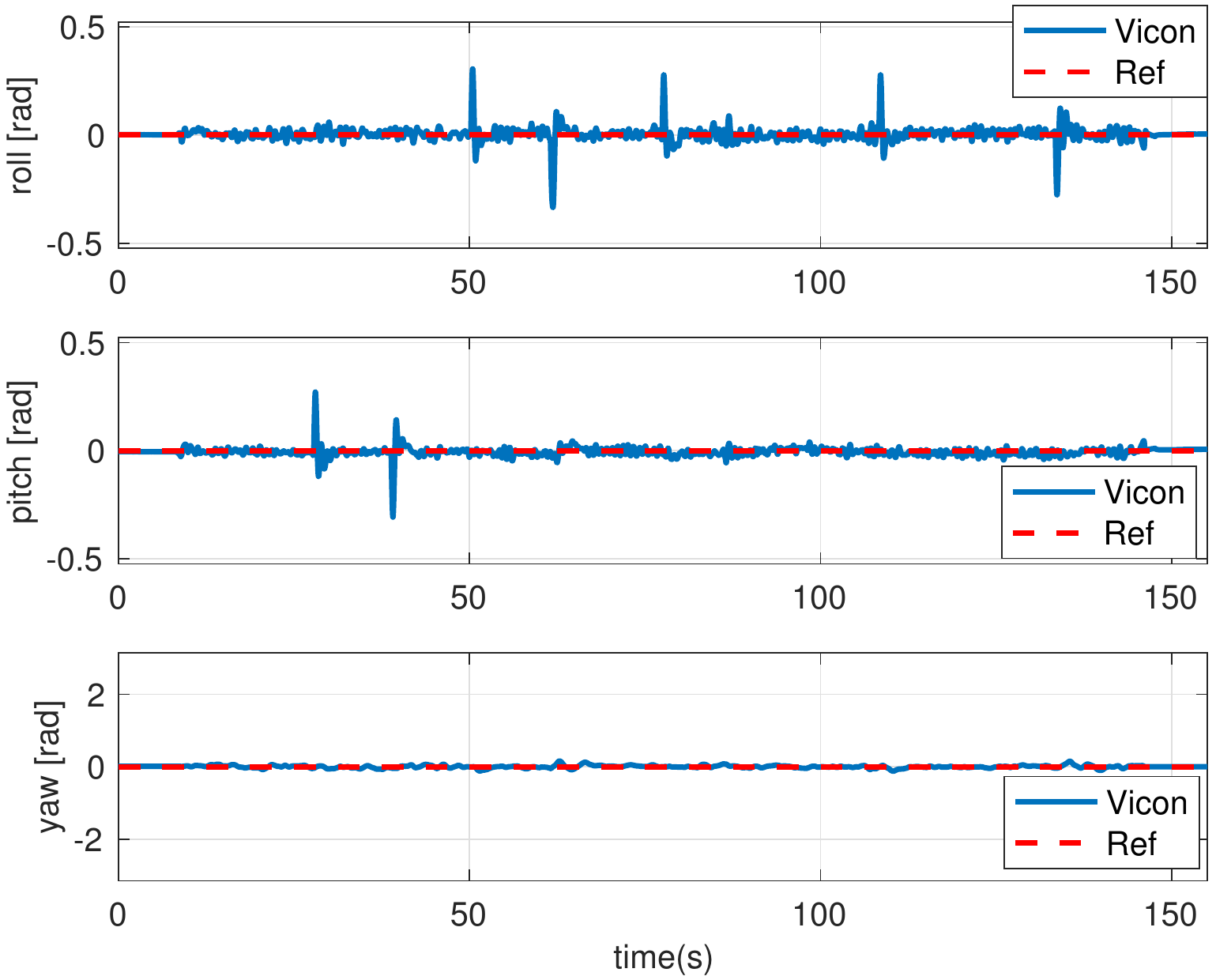}}
\centering
\vspace{-5mm}
\caption{Step response control performance without/with wind disturbances.}
\label{fig:stepResponse}
\vspace{-3mm}
\end{figure}

As oppose to the advantage of strong resistance to external disturbances, the downside for the heavy platform is a slower response. Fig.~\ref{fig:stepResponse} shows step response plots without wind disturbances (a) and (b), with wind (c) and (d). A goal position is manually chosen to excite all axises. The peaks in roll and pitch are caused by tilting toward the direction of maneuver, so we ignore them in RMS calculation. The results from Table~\ref{tbl:results_summary} show a large control error in both x and y. Slow response causes accumulating error while the vehicle reaches to a reference goal. Note that x RMS error in windy condition (0.127\unit{m}) is much smaller than without wind (0.259\unit{m}). It cannot be a fair comparison for the x state because the vehicle travels only 2\unit{m} along x-axis in windy condition in Fig.~\ref{fig:stepResponse}(c), whereas it moves 6\unit{m} in Fig.~\ref{fig:stepResponse}(a).

\subsubsection{Experimental results for trajectory following}
We use \cite{burri2015real} to generate a smooth polynomial reference trajectory as shown in Fig.~\ref{fig:3d_pose}. Even though hovering and step response tasks explicitly demonstrate control performance and essential functionalities for VTOL MAVs, trajectory following is also a significant task for many robotic applications such as obstacle avoidance, and path planning.

\begin{figure}
\centering
\subfigure[]{\includegraphics[width=0.49\columnwidth]{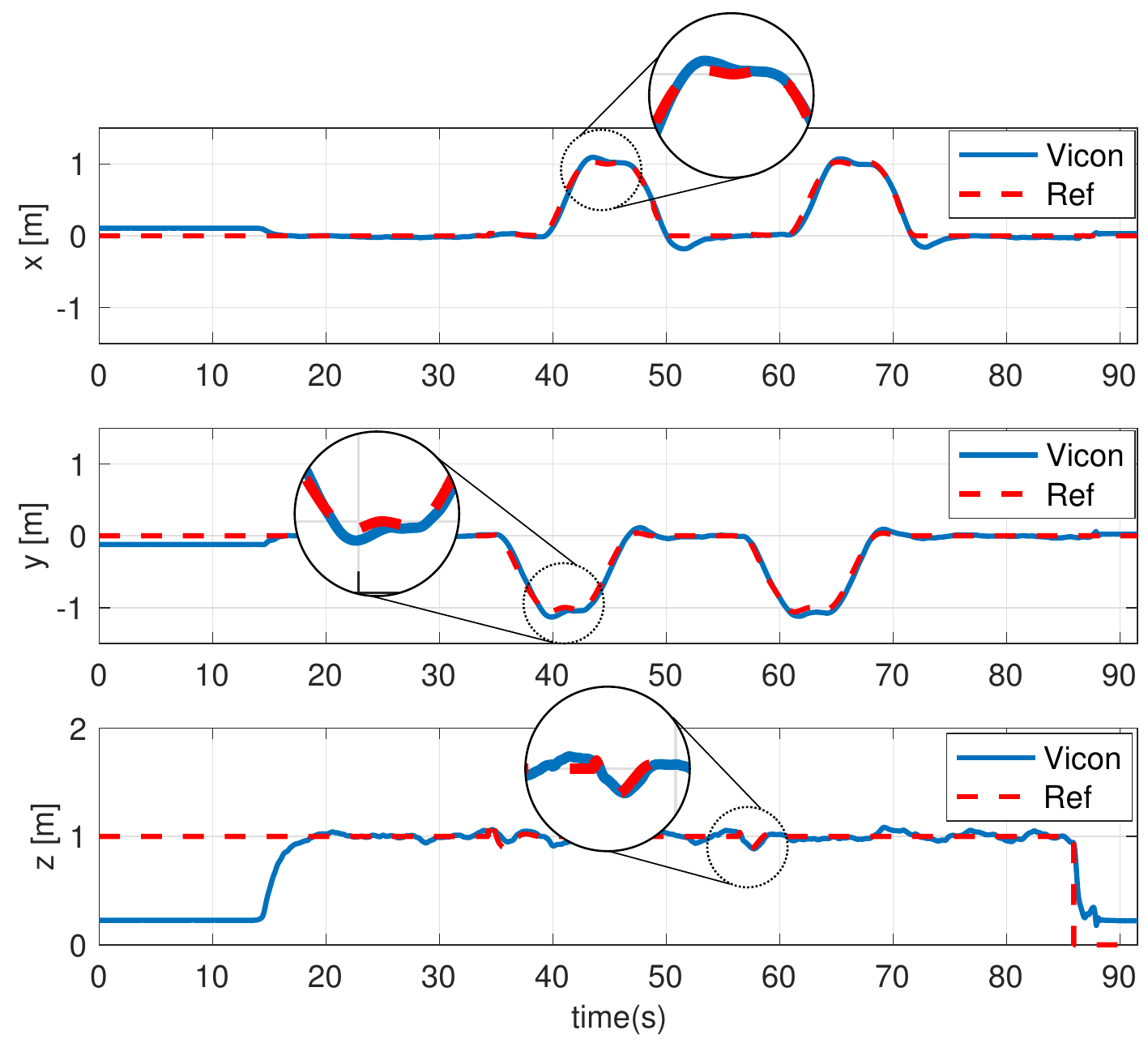}}
\subfigure[]{\includegraphics[width=0.49\columnwidth]{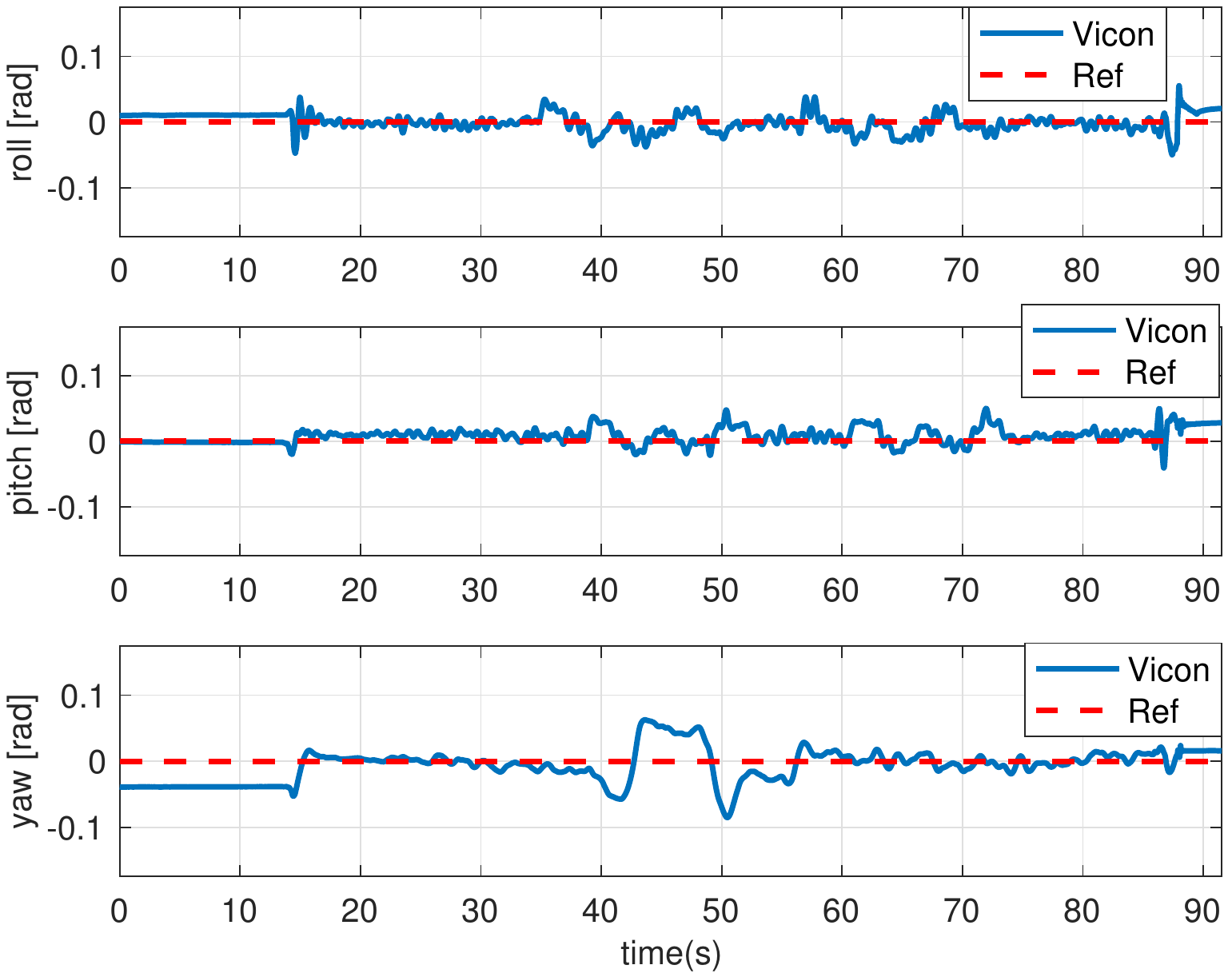}}
\subfigure[]{\includegraphics[width=0.49\columnwidth]{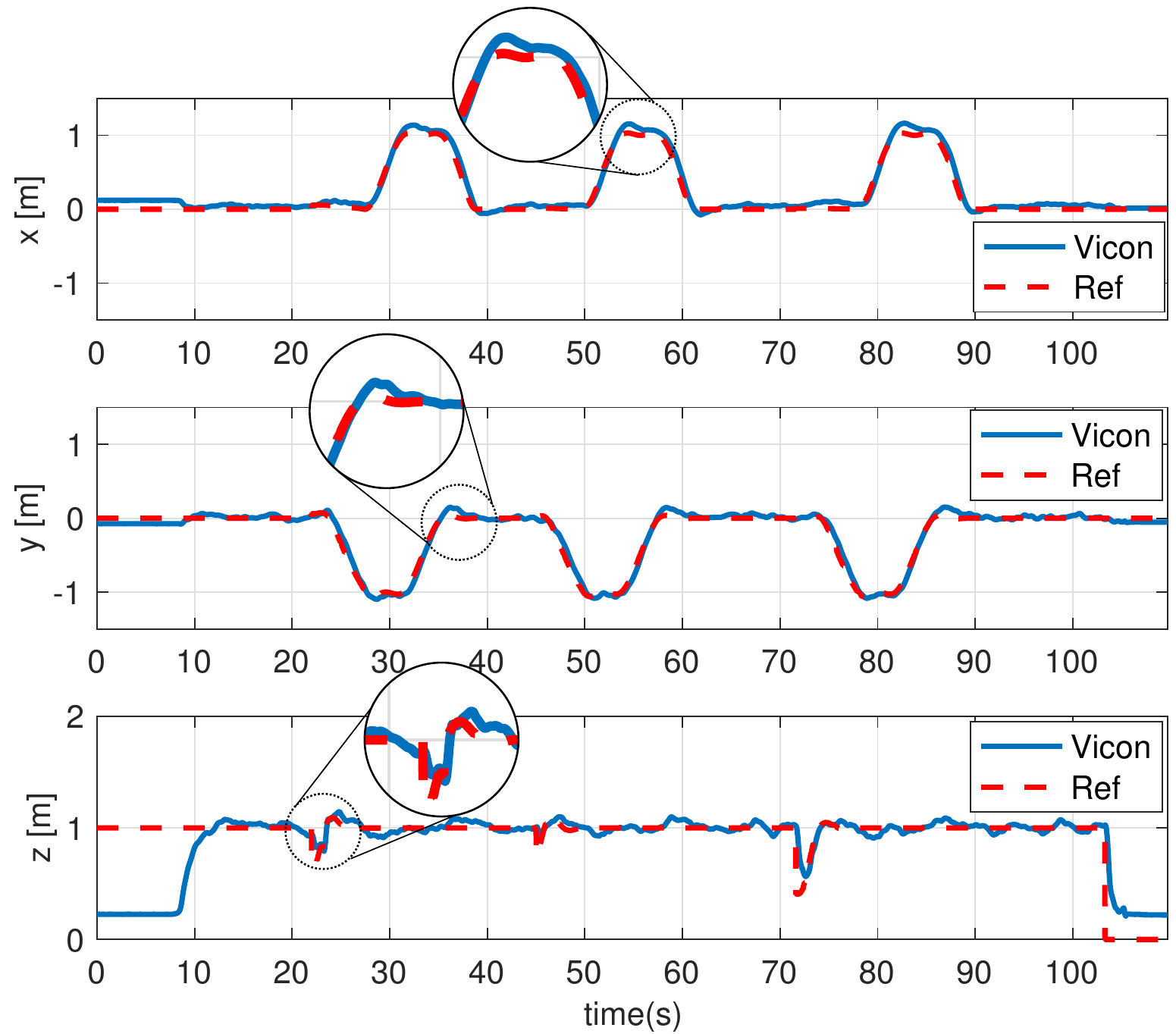}}
\subfigure[]{\includegraphics[width=0.49\columnwidth]{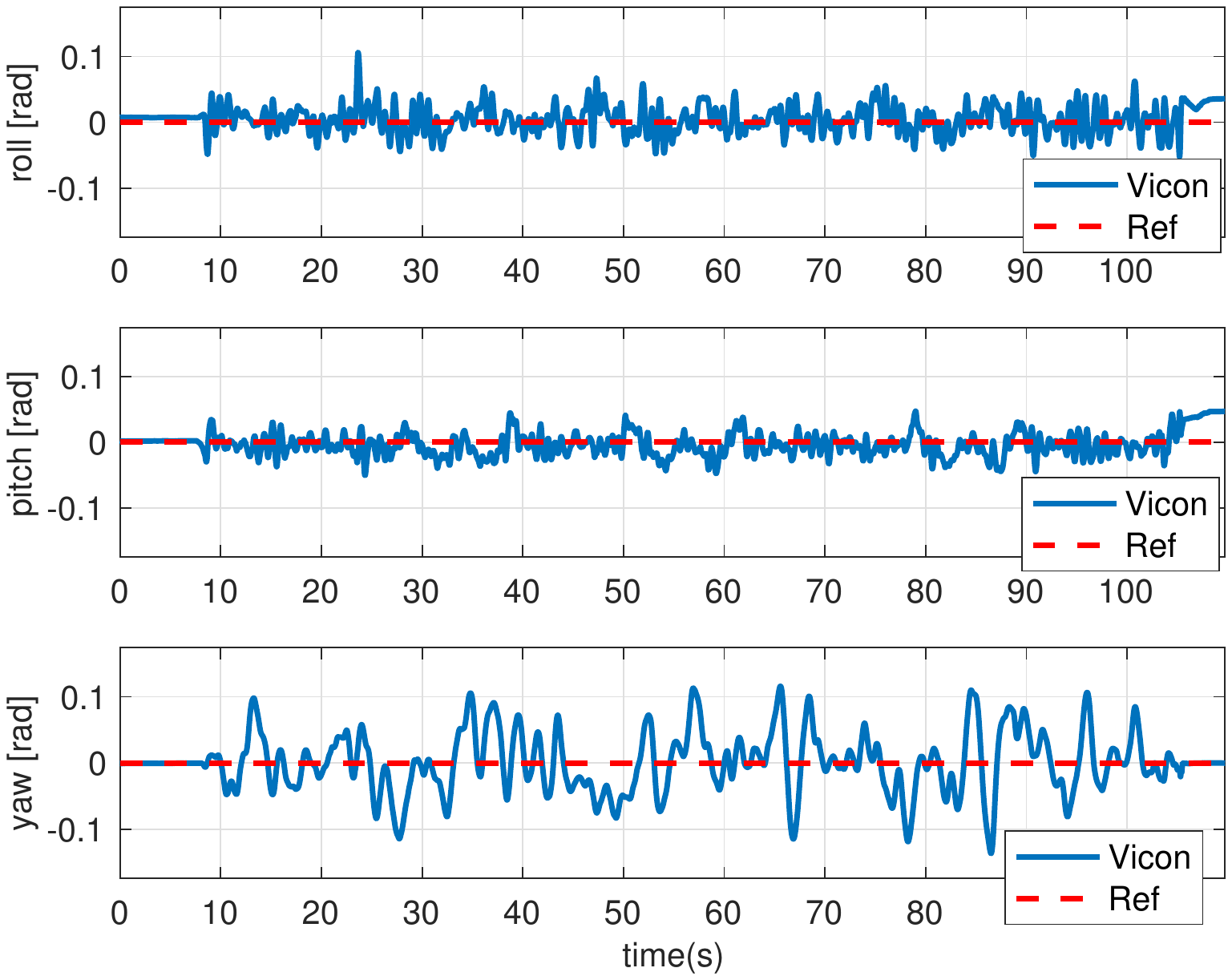}}
\centering
\vspace{-3mm}
\caption{Trajectory following control performance without/with wind disturbances.}
\label{fig:wp}
\vspace{-5mm}
\end{figure}

Fig.~\ref{fig:wp} shows trajectory following results without wind disturbances (a) and (b), and with the wind (c) and (d). It can be seen that the platform tracks the reference well in both conditions. There are noticeable error in yaw as shown in Fig.~\ref{fig:wp}(b) at around 40\unit{s}-50\unit{s}. These are caused by motion transitions from rolling (moving along the y-axis) to pitching (moving along the x-axis). During this short period, a small torque along the z-axis is generated and changes the heading direction. RMS errors are also presented in Table \ref{tbl:results_summary}.

\subsubsection{Qualitative results}
\begin{figure}
\centering
\subfigure{\includegraphics[width=0.49\columnwidth]{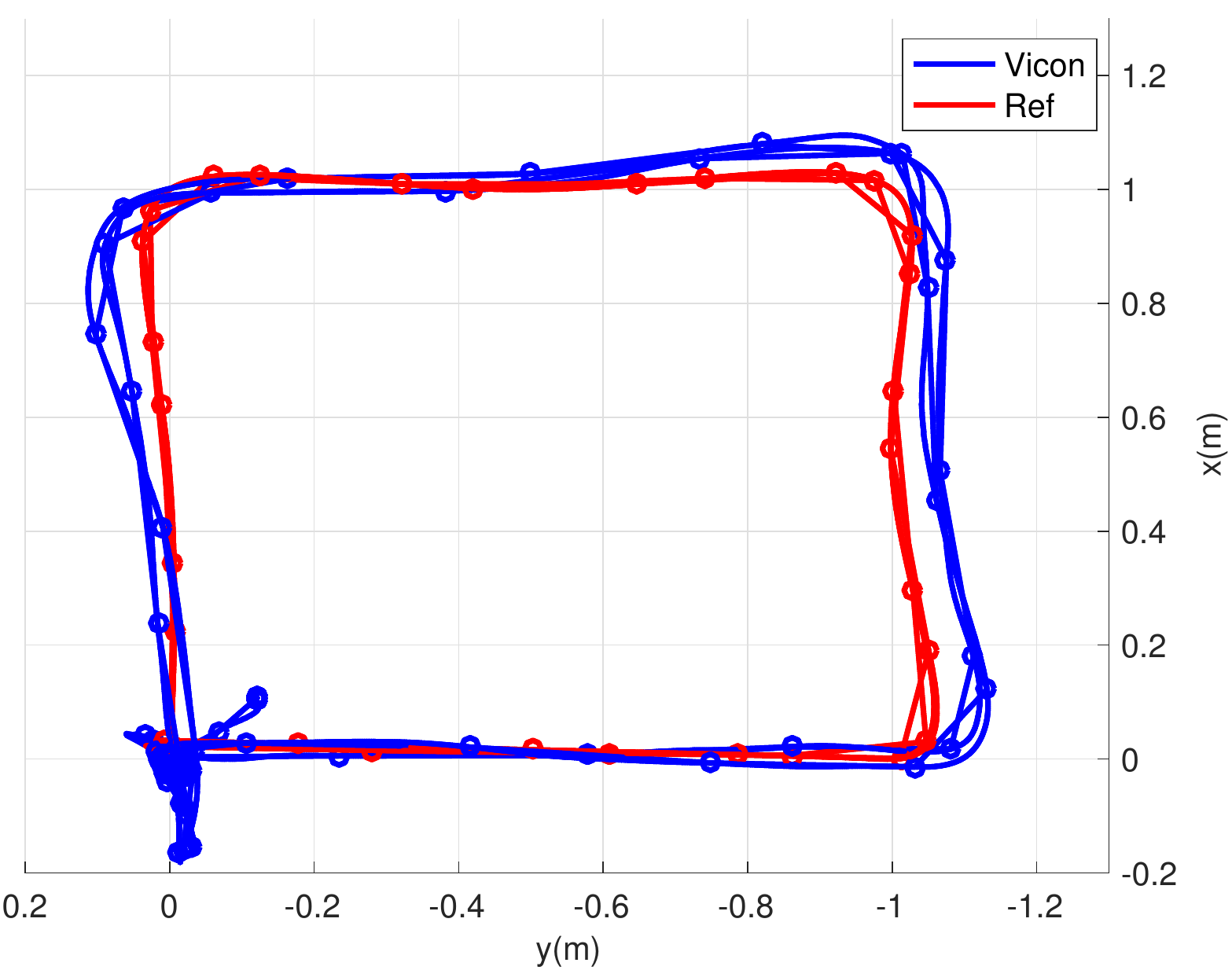}}
\subfigure{\includegraphics[width=0.49\columnwidth]{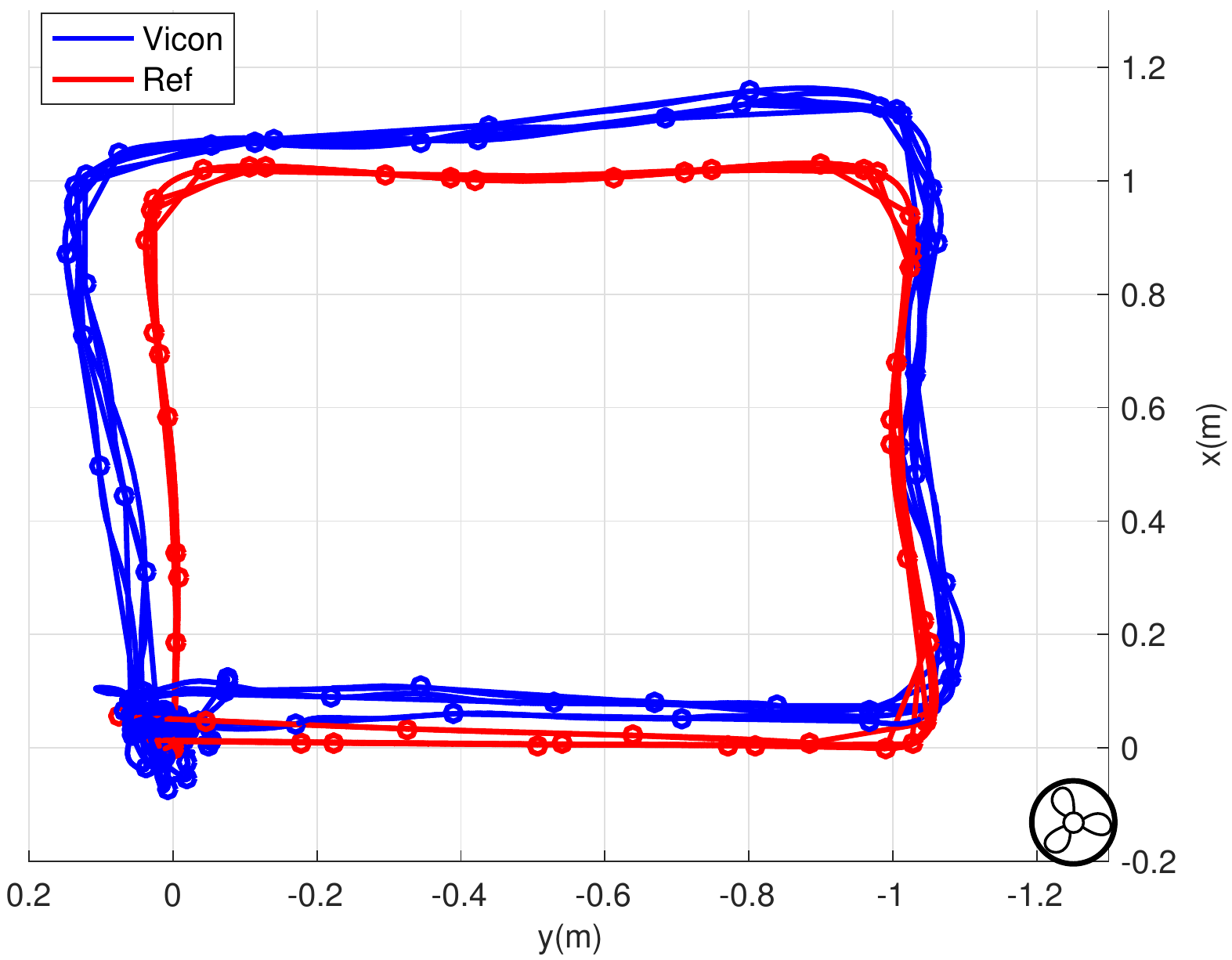}}
\subfigure{\includegraphics[width=0.49\columnwidth]{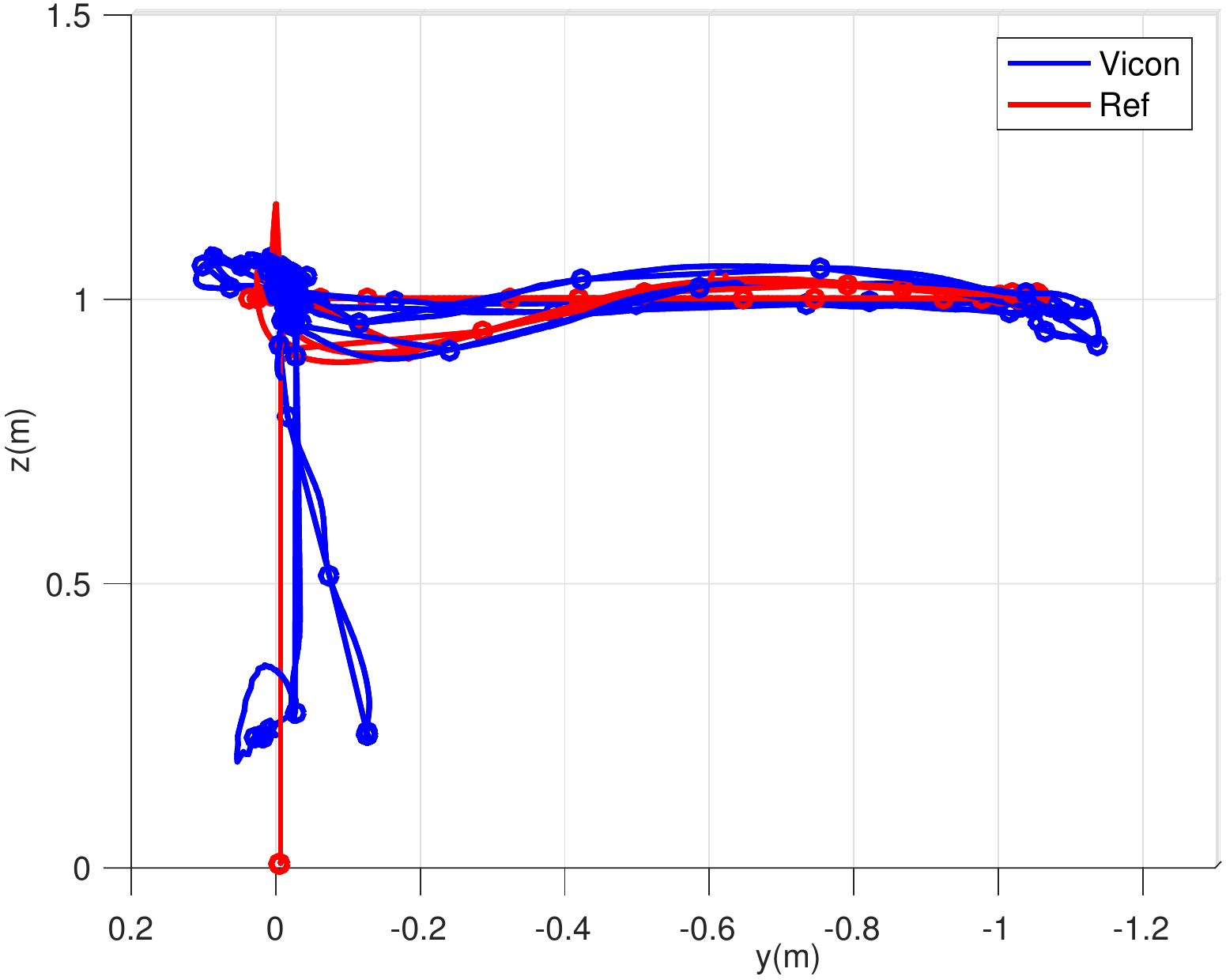}}
\subfigure{\includegraphics[width=0.49\columnwidth]{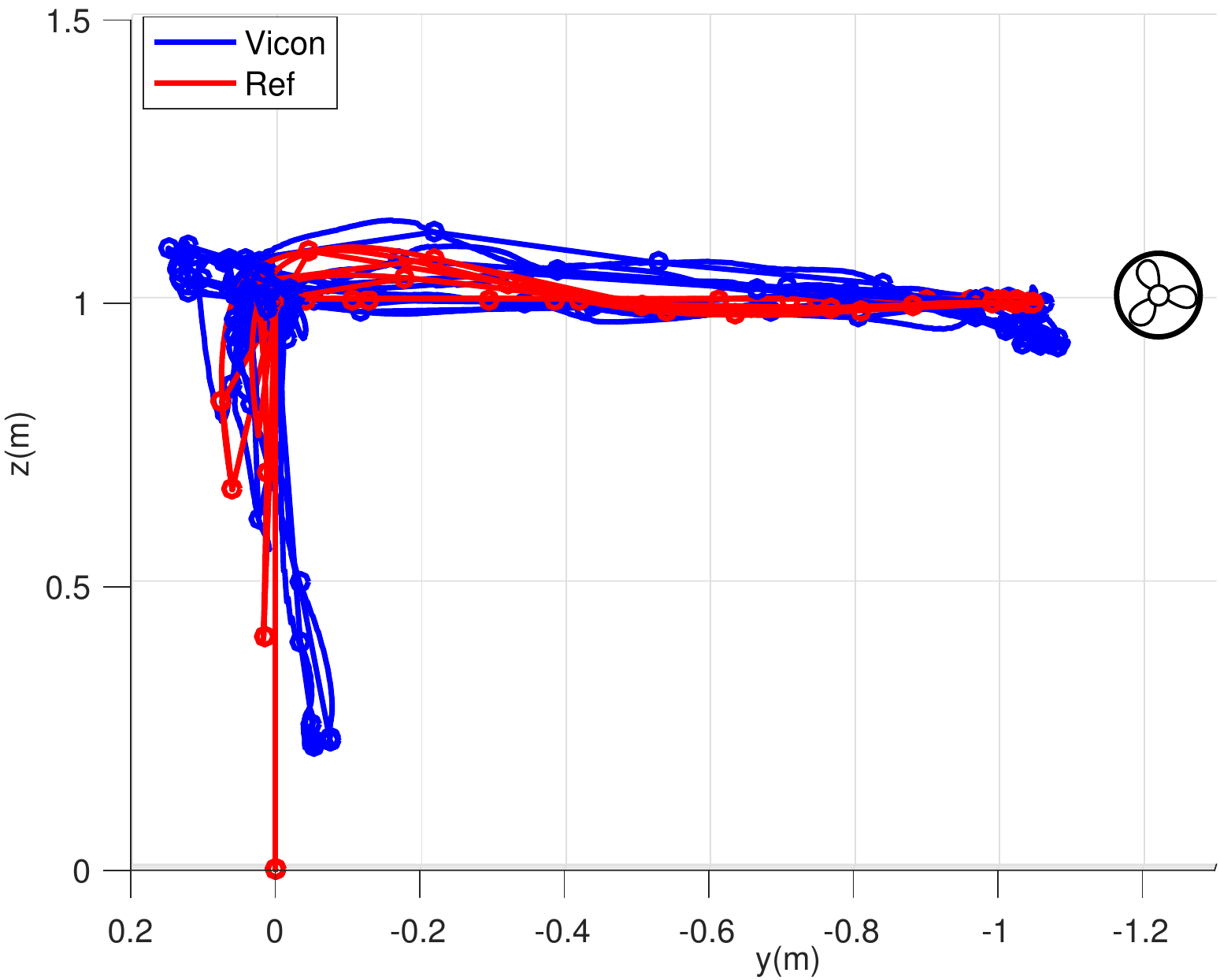}}
\subfigure{\includegraphics[width=0.49\columnwidth]{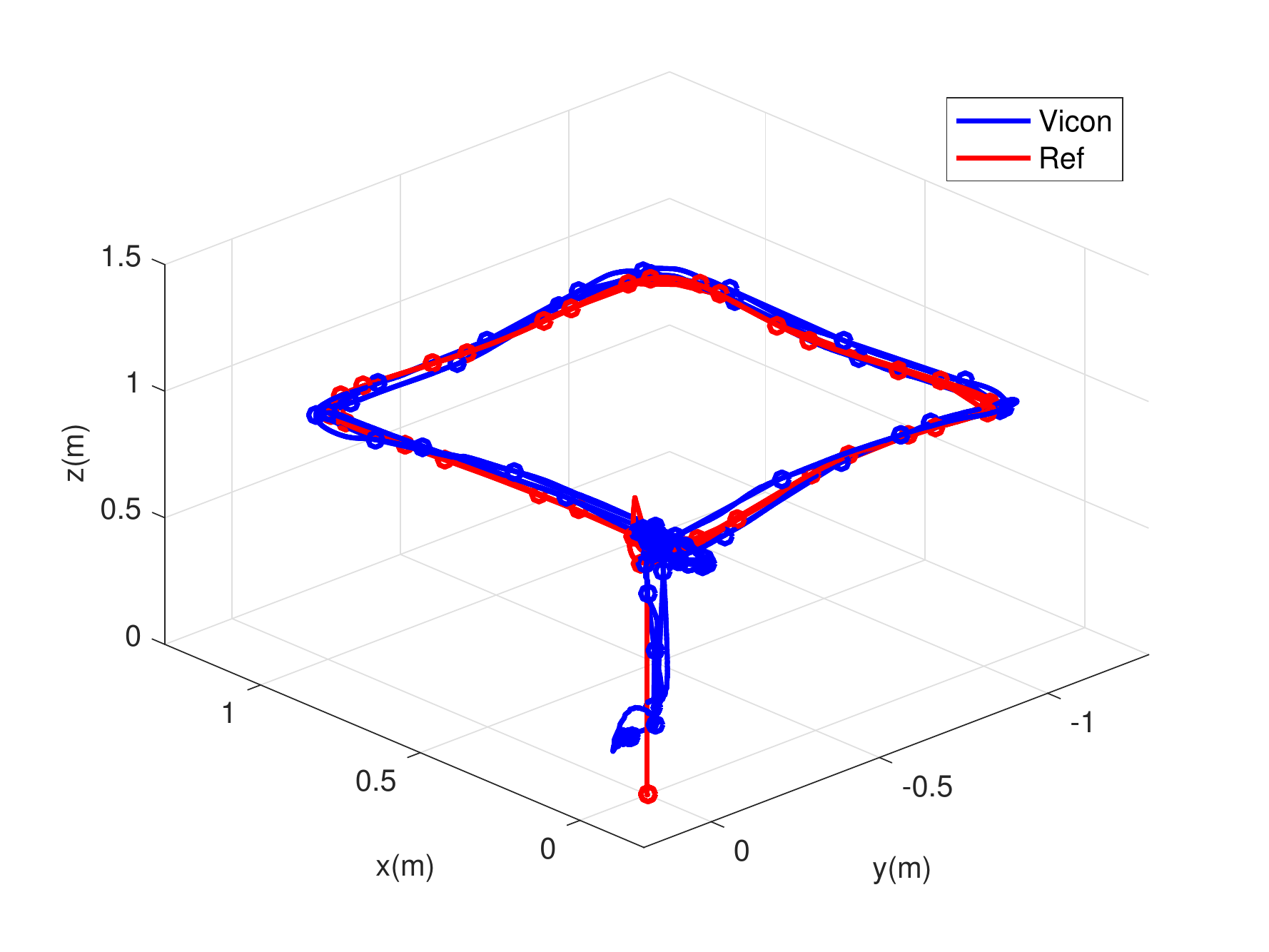}}
\subfigure{\includegraphics[width=0.49\columnwidth]{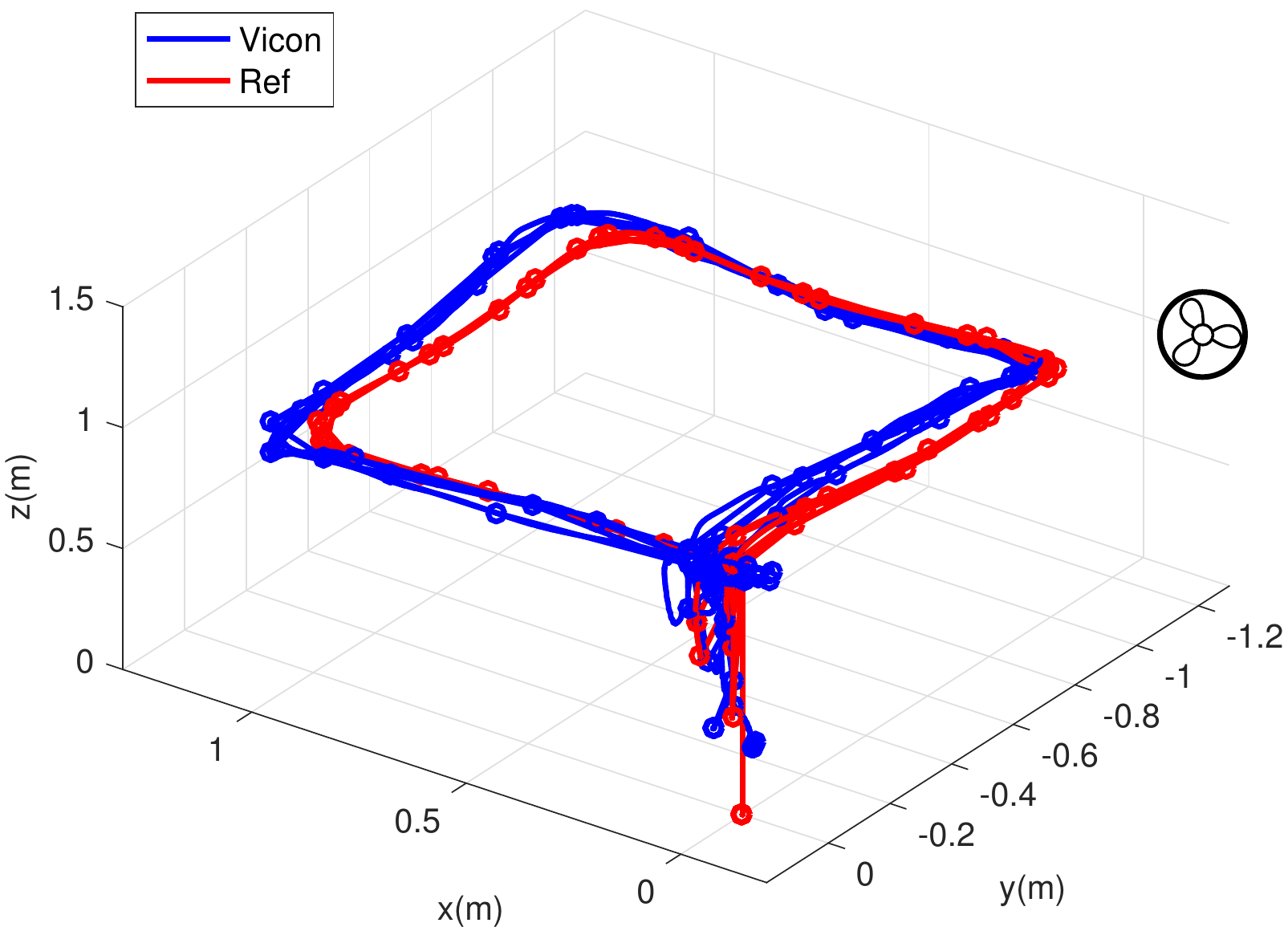}}
\centering
\caption{3D pose while trajectory following without (left column) and with (right column) wind disturbances.}
\label{fig:3d_pose}
\vspace{-2mm}
\end{figure}

\begin{figure}
\centering
\subfigure[]{\includegraphics[height=0.395\columnwidth]{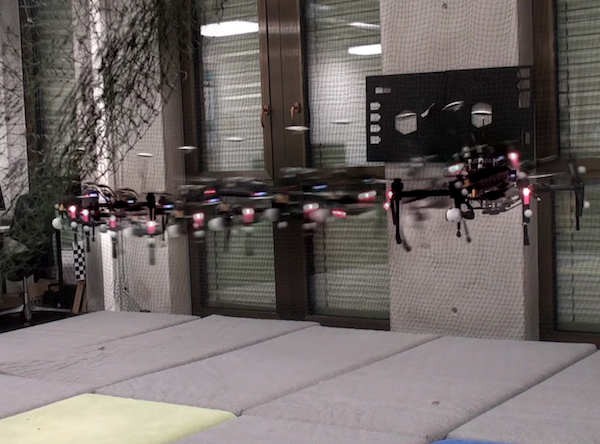}}
\subfigure[]{\includegraphics[height=0.395\columnwidth]{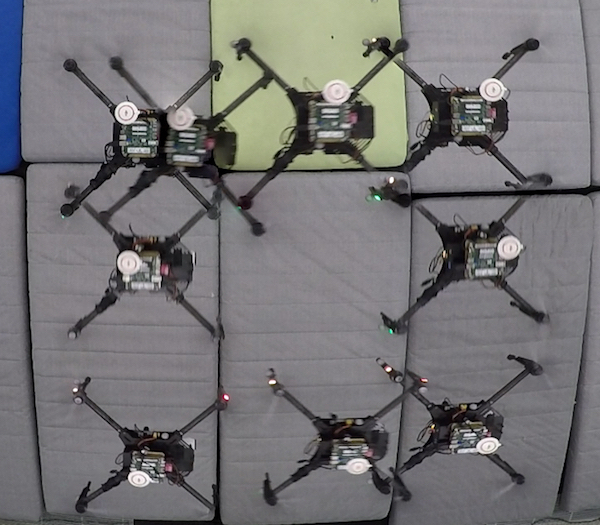}}
\centering
\vspace{-3mm}
\caption{Trajectory following control performance without (left column)/with wind disturbances (right column).}
\label{fig:time_laps}
\vspace{-5mm}
\end{figure}

We present two qualitative results for trajectory following and step response. Fig.~\ref{fig:3d_pose} illustrates the planned trajectory (red) and the vehicle position (blue) obtained from a motion capture device. The left column is without the wind and the right is in windy condition. Note that a fan is located around 3\unit{m} away from the hovering position along the South-East direction as illustrated. Each row is top, side, and perspective views. It is clearly seen that the trajectory is shifted to the wind direction (positive x and y-direction). Fig.~\ref{fig:time_laps} shows motions of step response (a) and trajectory following (b). For the step response, it can be seen that the vehicle builds up moments by tilting toward the goal direction and decelerates by tilting into the opposite direction when it approaches. For the trajectory following, the vehicle accurately follows 1$\times$1 square shown in Fig.~\ref{fig:time_laps}(b).

\section{Conclusions}\label{sec:conclusion}
\vspace{-2mm}
\addtolength{\textheight}{-3cm}   
We have presented good control performance of a cost-effect VTOL MAV platform using classic system identification techniques and the-state-of-the art MPC controller. The essential information for developing the VTOL MAV robotic platform such as attitude dynamics are addressed. The applied controller performance are evaluated in both simulation and experiments while executing hovering, step response and trajectory following. Using this platform has many advantages such its low-cost, high payload, ease of use and the ready availability of replacement parts, user friendly interface, powerful SDK, and a large user community. Our experiences and findings about the platform are returned to the community through open documentation and software packages to support researchers having their high-performance MAV platform for diverse robotic applications.

\bibliographystyle{ieeetr}
\bibliography{bibs/ICRA2017}

\begin{thebibliography}{10}

\bibitem{burri2015real}
M.~Burri, H.~Oleynikova, M.~W. Achtelik, and R.~Siegwart, ``Real-time
  visual-inertial mapping, re-localization and planning onboard mavs in unknown
  environments,'' in {\em Intelligent Robots and Systems (IROS), 2015 IEEE/RSJ
  International Conference on}, pp.~1872--1878, IEEE, 2015.

\bibitem{Nuske:2015aa}
S.~T. Nuske, S.~{Choudhury }, S.~Jain, A.~D. Chambers, L.~Yoder, S.~Scherer,
  L.~J. Chamberlain, H.~Cover, and S.~Singh, ``{Autonomous Exploration and
  Motion Planning for an Unmanned Aerial Vehicle Navigating Rivers},'' {\em
  Journal of Field Robotics}, June 2015.

\bibitem{Yang-:2015aa}
S.~{Yang }, Z.~Fang, S.~Jain, G.~{Dubey }, S.~M. Maeta, S.~Roth, S.~Scherer,
  Y.~Zhang, and S.~T. Nuske, ``{High-precision Autonomous Flight in Constrained
  Shipboard Environments},'' Tech. Rep. CMU-RI-TR-15-06, Robotics Institute,
  Pittsburgh, PA, February 2015.

\bibitem{Lee:2012aa}
D.~Lee, T.~Ryan, and H.~J. Kim, ``Autonomous landing of a vtol uav on a moving
  platform using image-based visual servoing,'' in {\em Robotics and Automation
  (ICRA), 2012 IEEE International Conference on}, pp.~971--976, IEEE, 2012.

\bibitem{mellinger2011design}
D.~Mellinger, Q.~Lindsey, M.~Shomin, and V.~Kumar, ``Design, modeling,
  estimation and control for aerial grasping and manipulation,'' in {\em 2011
  IEEE/RSJ International Conference on Intelligent Robots and Systems},
  pp.~2668--2673, IEEE, 2011.

\bibitem{Zhang:2012aa}
C.~Zhang and J.~M. Kovacs, ``The application of small unmanned aerial systems
  for precision agriculture: a review,'' {\em Precision agriculture}, vol.~13,
  no.~6, pp.~693--712, 2012.

\bibitem{Achtelik:2011fk}
M.~Achtelik, S.~Weiss, and R.~Siegwar, ``{Onboard IMU and Monocular Vision
  Based Control for MAVs in Unknown In- and Outdoor Environments},'' in {\em
  Proceedings of the IEEE International Conference on Robotics and Automation
  (ICRA)}, May 2011.

\bibitem{Weiss:2011aa}
S.~Weiss, D.~Scaramuzza, and R.~Siegwart, ``Monocular-slam--based navigation
  for autonomous micro helicopters in gps-denied environments,'' {\em Journal
  of Field Robotics}, vol.~28, no.~6, pp.~854--874, 2011.

\bibitem{bouabdallah2007design}
S.~Bouabdallah, {\em Design and control of quadrotors with application to
  autonomous flying}.
\newblock PhD thesis, Ecole Polytechnique Federale de Lausanne, 2007.

\bibitem{mahony2012multirotor}
R.~Mahony, V.~Kumar, and P.~Corke, ``{Multirotor aerial vehicles: Modeling,
  estimation, and control of quadrotor},'' {\em IEEE Robotics and Automation
  Magazine}, vol.~19, no.~3, pp.~20--32, 2012.

\bibitem{Pounds:2009fk}
P.~Pounds and R.~Mahony, ``Design principles of large quadrotors for practical
  applications,'' in {\em 2009 IEEE International Conference on Robotics and
  Automation (ICRA)}, May 2009.

\bibitem{hoffmann2008quadrotor}
G.~M. Hoffmann, S.~L. Waslander, and C.~J. Tomlin, ``Quadrotor helicopter
  trajectory tracking control,'' in {\em AIAA guidance, navigation and control
  conference and exhibit}, pp.~1--14, 2008.

\bibitem{Pounds:2010aa}
P.~Pounds, R.~Mahony, and P.~Corke, ``Modelling and control of a large
  quadrotor robot,'' {\em Control Engineering Practice}, vol.~18, no.~7,
  pp.~691--699, 2010.

\bibitem{chowdhary2010aerodynamic}
G.~Chowdhary and R.~Jategaonkar, ``Aerodynamic parameter estimation from flight
  data applying extended and unscented kalman filter,'' {\em Aerospace science
  and technology}, vol.~14, no.~2, pp.~106--117, 2010.

\bibitem{Sa:2012ICRA}
I.~Sa and P.~Corke, ``{System Identification, Estimation and Control for a Cost
  Effective Open-Source Quadcopter},'' in {\em 2012 IEEE International
  Conference on Robotics and Automation (ICRA)}, 2012.

\bibitem{tischler2006aircraft}
M.~B. Tischler and R.~K. Remple, ``Aircraft and rotorcraft system
  identification,'' {\em AIAA education series}, 2006.

\bibitem{burri2016maximum}
M.~Burri, J.~Nikolic, H.~Oleynikova, M.~W. Achtelik, and R.~Siegwart,
  ``{Maximum likelihood parameter identification for MAVs},'' in {\em 2016 IEEE
  International Conference on Robotics and Automation (ICRA)}, IEEE, 2016.

\bibitem{kamelmpc2016}
M.~Kamel, T.~Stastny, K.~Alexis, and R.~Siegwart, ``{Model Predictive Control
  for Trajectory Tracking of Unmanned Aerial Vehicles Using Robot Operating
  System},'' in {\em Robot Operating System (ROS) The Complete Reference}
  (A.~Koubaa, ed.), Springer Press, (to appear) 2016.

\bibitem{FORCESPro}
A.~Domahidi and J.~Jerez, ``{FORCES Professional}.'' {embotech GmbH
  (\nobreak{\url{http://embotech.com/FORCES-Pro}})}, July 2014.

\bibitem{Quigley:2009aa}
M.~Quigley, K.~Conley, B.~Gerkey, J.~Faust, T.~Foote, J.~Leibs, R.~Wheeler, and
  A.~Y. Ng, ``{ROS: an open-source Robot Operating System},'' in {\em ICRA
  workshop on open source software}, vol.~3, p.~5, Kobe, Japan, 2009.

\end{thebibliography}

\end{document}